\documentclass{article}

\usepackage[preprint]{neurips_2026}

\usepackage[utf8]{inputenc} 
\usepackage[T1]{fontenc}    
\usepackage{booktabs}       
\usepackage{amsmath,amssymb,amsfonts}
\usepackage{mathrsfs}
\usepackage{nicefrac}       
\usepackage{microtype}      
\usepackage{xcolor}
\usepackage{graphicx}
\usepackage{multirow}
\usepackage{textcomp}
\usepackage{siunitx}        
\usepackage{algorithm}
\usepackage{algorithmicx}
\usepackage{algpseudocode}
\usepackage{url}            
\usepackage{hyperref}       

\graphicspath{{./figures/}}

\title{Neuromorphic Energy-Aware Learning for Adaptive Deep Brain Stimulation}

\author{%
  \begin{minipage}{\dimexpr\textwidth-2\tabcolsep\relax}
    \centering\normalfont
    {\bfseries
      Binh Nguyen\textnormal{\textsuperscript{1}}\quad
      Colleen Josephson\textnormal{\textsuperscript{1}}\quad
      Mircea Teodorescu\textnormal{\textsuperscript{1}}\\[3pt]
      Gert Cauwenberghs\textnormal{\textsuperscript{2}}\quad
      Jason Eshraghian\textnormal{\textsuperscript{1}}%
    }\thanks{Corresponding author: \texttt{jeshragh@ucsc.edu}}\\[5pt]
    {\normalfont
      \textsuperscript{1}Dept.\ of Electrical and Computer Engineering, University of California, Santa Cruz, CA, USA\\
      \textsuperscript{2}Dept.\ of Bioengineering, University of California, San Diego, La Jolla, CA, USA%
    }
  \end{minipage}%
}

\begin{document}

\maketitle

\begin{abstract}
  Neuromorphic and edge computing research has focused on reducing the inference cost of neural network controllers, yet in physical closed-loop systems the actuator can rival or exceed an efficient controller in energy. An efficient controller is therefore necessary but not sufficient, because the actuator becomes the cost worth reducing once inference no longer dominates it. Here, we introduce energy-aware learning, an approach that incorporates actuator energy directly into the reinforcement learning reward, and demonstrate it in closed-loop deep brain stimulation (DBS) for Parkinson's disease. A deep spiking Q-network, trained in a biophysical cortico-basal ganglia-thalamic circuit model, learns to suppress pathological alpha-beta oscillations by 45.2\% while reducing stimulation charge by 80.0\% relative to continuous DBS. Sparsity-constrained knowledge distillation compresses the policy onto the SynSense XyloAudio~3 neuromorphic processor at 0.52~mW inference power, yielding 28.1$\times$ lower energy per inference than an equivalent artificial neural network on conventional edge hardware. By co-optimizing stimulation energy and inference efficiency, the framework addresses both major power demands in implantable neuromodulation.
\end{abstract}

\section{Introduction}\label{sec1}
The deployment of neural networks on edge and neuromorphic architectures has achieved substantial reductions in computational power. However, when these networks operate as controllers in physical closed-loop systems, the energy expended by the actuator is a first-order cost, comparable to or exceeding the inference itself once the controller runs on efficient hardware. In deep brain stimulation (DBS), the neurostimulator's continuous charge delivery draws power on the order of hundreds of microwatts to a milliwatt~\cite{saricaImplantablePulseGenerators2021, helmersComparisonBatteryLife2018}, the same regime as a sub-milliwatt neuromorphic controller. Reducing inference alone therefore addresses only part of the implant’s energy budget. This observation motivates energy-aware learning: incorporating the output energy of the controlled system directly into the learning objective, so that the policy optimizes what it can practically reduce.

We demonstrate this principle in closed-loop DBS for Parkinson's Disease (PD). DBS is the primary surgical therapy for advanced PD, reducing tremor, rigidity, and bradykinesia~\cite{benabidDeepBrainStimulation2009}, yet conventional devices deliver continuous (cDBS), high-frequency electrical pulses regardless of clinical state, leading to stimulation-induced side effects (e.g., dysarthria, gait disturbance) and wasteful battery usage~\cite{habetsUpdateAdaptiveDeep2018}. In non-rechargeable Implantable Pulse Generators (IPGs), high duty cycles necessitate frequent surgical replacements, with pulse width exhibiting a negative correlation with time-to-replacement~\cite{sharmaStudyBatteryReplacement2023}. These limitations have driven interest in adaptive DBS (aDBS), which modulates stimulation parameters in real-time based on biomarkers of pathological network activity~\cite{arlottiEighthoursAdaptiveDeep2018, velisarDualThresholdNeural2019}. Because stimulation accounts for a substantial fraction of an IPG's power budget once inference is efficient, an effective aDBS controller must optimize stimulation energy alongside therapeutic efficacy, not merely minimize its own inference cost.

\begin{figure}[t!]
  \centering
  \includegraphics[width=\linewidth]{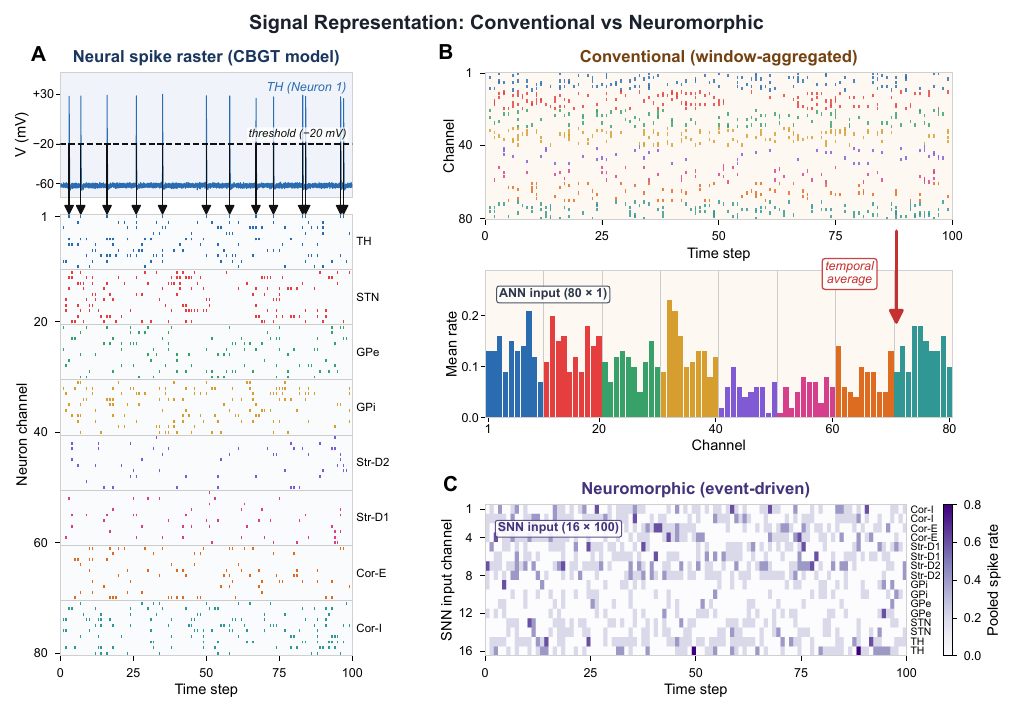}
  \caption{\textbf{Signal representation: conventional versus neuromorphic.}
    \textbf{(A)}~Raw brain signals from the CBGT model. Top: A sample continuous membrane potential trace demonstrating spike generation via threshold crossing ($-20$~mV). Bottom: The resulting 80-channel spike raster across eight basal ganglia populations (TH, STN, GPe, GPi, Str-D2, Str-D1, Cor-E, Cor-I).
    \textbf{(B)}~Conventional window-aggregated processing collapses the temporal axis via mean pooling, reducing the full spike raster to an 80-dimensional rate vector (ANN input: $80 \times 1$). All spike-timing information, including inter-burst synchrony that encodes pathological state, is discarded.
  \textbf{(C)}~Neuromorphic event-driven encoding compresses the spatial axis (80 $\to$ 16 channels via average pooling) while preserving the full 100-step temporal sequence (SNN input: $16 \times 100$). The SNN processes this time series natively through its membrane dynamics, retaining cross-population synchrony patterns without requiring explicit recurrent layers.}
  \label{fig:signal_comparison}
\end{figure}

The most prominent biomarker for PD is the exaggerated $\alpha$-$\beta$ band (7--35~Hz; also referred to as the $\beta$-band or simply beta throughout) oscillation observed in the subthalamic nucleus (STN) and globus pallidus internus (GPi)~\cite{hammondPathologicalSynchronizationParkinsons2007, littleWhatBrainSignals2012}. Prior work has applied neural network controllers to model-based closed-loop DBS, achieving energy-efficient beta suppression in computational Cortico-Basal Ganglia-Thalamocortical (CBGT) models~\cite{liuNeuralNetworkBasedClosedLoop2020, gaoModelBasedDesignClosed2020, choClosedLoopDeepBrain2024, luApplicationReinforcementLearning2020, suModelBasedEvaluationClosedLoop2019}. However, these approaches rely on conventional hardware and window-aggregated processing, requiring aggregation of neural data into static windowed features (e.g., average spectral power) and discarding the fine-grained temporal structure of the underlying spike train. Figure~\ref{fig:signal_comparison} illustrates that the raw CBGT signals, bridging continuous membrane potentials to discrete spike events (Fig.~\ref{fig:signal_comparison}A), encode precise firing times across eight basal ganglia populations, whereas conventional processing reduces this to per-window mean rates (Fig.~\ref{fig:signal_comparison}B), discarding the inter-burst timing that tracks pathological synchrony. Deploying such controllers on IPGs is further constrained by tissue-heating safety limits---the implant must not raise the temperature of surrounding tissue by more than $\sim$2$^\circ$C~\cite{parastarfeizabadiAdvancesClosedloopDeep2017}---and by the need for multi-year battery life. Together these constraints restrict an always-on controller to a sub-milliwatt power envelope, orders of magnitude below the watt-scale draw of conventional edge AI accelerators~\cite{barchiEnergyEfficientLowlatency2024}.

Fortunately, neuromorphic computing operates natively in this constrained regime. Spiking Neural Networks (SNNs), which mimic the event-driven processing of biological neurons, can process neural signals at sub-milliwatt power and sub-millisecond latency~\cite{thakurLargeScaleNeuromorphicSpiking2018}. Unlike conventional artificial neural networks (ANNs) and recurrent neural networks (RNNs), SNNs compute only when input spikes arrive and are able to exploit internal delays to capture spatio-temporal dependencies~\cite{meszarosEfficientEventbasedDelay2025}, making them well matched to the asynchronous nature of biological signals. This encoding preserves temporal relationships in a pooled, event-driven format that directly feeds into an SNN controller (Fig.~\ref{fig:signal_comparison}C).

\begin{figure}[t!]
  \centering
  \includegraphics[width=\linewidth]{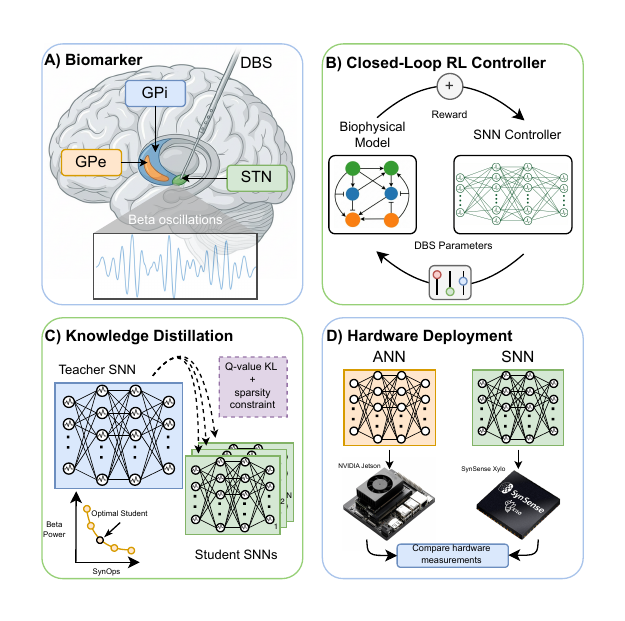}
  \caption{\textbf{Neuromorphic closed-loop DBS framework.}
    \textbf{(A)}~Pathological $\beta$ oscillations in the parkinsonian basal ganglia circuit targeted by DBS.
    \textbf{(B)}~A spiking RL controller (DSQN) modulates DBS parameters in a closed loop with a biophysical CBGT model.
    \textbf{(C)}~Sparsity-constrained distillation compresses the teacher SNN into sparse students, trading synaptic operations (SynOps) for therapeutic efficacy.
  \textbf{(D)}~Hardware deployment of the distilled SNN on the SynSense XyloAudio~3 is compared against an equivalent ANN on the NVIDIA Jetson Orin Nano to quantify power, latency, and energy per inference.}
  \label{fig:system_overview}
\end{figure}

Here, we present an end-to-end neuromorphic closed-loop DBS system that processes neural activity in its native spiking domain (Fig.~\ref{fig:system_overview}). We train a Deep Spiking Q-Network (DSQN)~\cite{nguyenAcceleratingNeuromorphicDeepBrain2025, nguyenClosedLoopNeuromorphicDeep2025} whose reward function jointly optimizes therapeutic $\beta$ suppression and the physical charge delivered by the neurostimulator, then deploy the resulting policy on neuromorphic hardware via sparsity-constrained knowledge distillation.

Our contributions span three areas:
\begin{itemize}
  \item \textbf{Energy-Aware Control:} The reinforcement learning (RL) reward function directly penalizes stimulation charge delivery, targeting a first-order power demand in the implant. At 0.52~mW inference power, the neuromorphic controller operates within the sub-milliwatt envelope required by strict thermal and multi-year battery constraints, ensuring the controller itself does not offset the stimulation savings.
  \item \textbf{Neuromorphic Benchmarking:} Deploying the SNN on the SynSense XyloAudio 3 neuromorphic processor~\cite{bosSubmWNeuromorphicSNN2022} and benchmarking against ANN and RNN baselines on an NVIDIA Jetson Orin Nano~\cite{nvidiaJetsonNano2019}, we achieve a 28.1$\times$ reduction in energy per inference while quantifying power and latency trade-offs.
  \item \textbf{Biophysical Validation:} We validate all control policies in simulation against Hodgkin-Huxley~\cite{hodgkinQuantitativeDescriptionMembrane1952} dynamics in the 6-hydroxydopamine (6-OHDA) lesioned rat model of the CBGT network~\cite{kumaraveluBiophysicalModelCortexbasal2016}, achieving 45.2\% pathological $\alpha$-$\beta$ band-power reduction (offline multi-taper spectral metric) and an 80.0\% reduction in cumulative stimulation charge relative to continuous DBS.
\end{itemize}

\section{Methods}\label{sec4}
\subsection{Biophysical Simulation Environment}
We validated control policies in a biophysically detailed simulation of the rat CBGT circuit based on Kumaravelu et al.~\cite{kumaraveluBiophysicalModelCortexbasal2016}. The network comprises 80 single-compartment neurons across eight populations: cortical excitatory and inhibitory (Izhikevich dynamics~\cite{izhikevichSimpleModelSpiking2003}), and subthalamic nucleus (STN), globus pallidus externus (GPe), globus pallidus internus (GPi), striatum (D1 and D2 medium spiny neurons), and thalamus (TH), all modeled with conductance-based Hodgkin-Huxley dynamics~\cite{hodgkinQuantitativeDescriptionMembrane1952}. Each subcortical population possesses nucleus-specific ionic currents (e.g., L-type and T-type calcium, A-type potassium, and calcium-dependent potassium currents in STN) that produce characteristic firing patterns including pacemaking and bursting. Full membrane equations, gating kinetics, and parameter tables are provided in Appendices~\ref{app:dynamics} and~\ref{app:netparams}.

The dopamine-depleted parkinsonian state was induced by reducing the maximal conductance of the striatal M-current ($g_M$) and corticostriatal synaptic strength ($g_{costr}$) while increasing intra-GPe inhibition ($g_{GPe-GPe}$)~\cite{kumaraveluBiophysicalModelCortexbasal2016}, reproducing the exaggerated $\alpha$-$\beta$ band (7--35~Hz) oscillations characteristic of PD~\cite{habetsUpdateAdaptiveDeep2018}. To ensure robustness to biological variability, network connectivity and initial conditions were randomized for each training episode. Inter-population synapses use conductance-based kinetics with bi-exponential rise and decay, introducing realistic temporal latencies (e.g., NMDA $\tau_{decay} \approx 67$--$90$~ms) that the controller must navigate (see Appendix~\ref{app:dynamics} for synaptic equations). Equations are integrated using Forward Euler with $dt = 0.01$~ms.

Spike events are registered as rising-phase threshold crossings at $V_{\mathrm{thr}} = -20$~mV, producing an 80-channel binary spike matrix (8 populations $\times$ 10 neurons) that constitutes the controller's state observation at each decision step. GPi $\alpha$-$\beta$ power is computed with a multi-taper point-process estimator using discrete prolate spheroidal sequences (DPSS) over per-100~ms analysis windows, applied consistently across all band-power quantification in this work~\cite{jarvisSamplingPropertiesSpectrum2001} (see Appendix~\ref{app:dynamics}). The same estimator provides the online biomarker that the simulator feeds to the RL reward and the proxy used to rank candidates in the Pareto front sweep (Fig.~\ref{fig:pareto}), where therapeutic efficacy is the mean of per-seed reductions across ten seeds, matching the protocol used for Fig.~\ref{fig:efficacy}. In both evaluations the teacher and all student models receive the spatially-pooled 16-channel representation, and all downstream policy evaluation uses this as input from the biophysical CBGT model's 80-channel spike matrix. This 4-second integrated $\alpha$-$\beta$ power is the reported metric in Fig.~\ref{fig:efficacy} (therapeutic efficacy) and the efficacy axis of Fig.~\ref{fig:pareto} (energy--efficacy trade-off). For paired comparisons of suppression across the $n=10$ seeds, we report Cohen's $d_z$ as the effect size, computed as the mean of the per-seed differences divided by their standard deviation.

\subsection{Deep Spiking Q-Network Architecture}
A DSQN is implemented to select discrete stimulation parameters, modulating the current driven by the stimulator. A Q-network is a value-based reinforcement learning architecture that estimates the expected future cumulative reward (Q-value) associated with taking a specific action in a given state. We specifically selected this off-policy, discrete-action approach over continuous policy gradient methods (e.g., Proximal Policy Optimization or Actor-Critic) for three primary reasons: (i) clinical DBS neurostimulators inherently operate using discrete, safety-bounded parameter step adjustments rather than continuous analog values, making Q-learning structurally aligned with the hardware interface; (ii) its off-policy nature permits the use of an experience replay buffer, which substantially improves sample efficiency when training against computationally expensive biophysical simulations; and (iii) the deterministic argmax selection of maximum Q-values maps natively to a competitive winner-take-all (WTA) readout layer, an optimal decoding strategy for event-driven spiking neuromorphic cores. Simultaneous co-modulation of these variables governs both the side-effect profile~\cite{donatiNeuromorphicHardwareSomatosensory2024} and battery longevity~\cite{sharmaStudyBatteryReplacement2023}. The network is implemented in PyTorch~\cite{paszke2019pytorch} using the Rockpool neuromorphic library~\cite{muir_dylan_2019_4639684}, which provides differentiable dynamics suitable for gradient-based optimization via surrogate gradients~\cite{eshraghianTrainingSpikingNeural2023a}. The SNN is constructed using Leaky Integrate-and-Fire (LIF) neurons with explicit synaptic current dynamics. In the discrete-time forward Euler formulation used by Rockpool, the synaptic current $I_s$ and membrane potential $V_m$ evolve as a coupled system:
\begin{align}
  I_{s}^{t+1} &= I_{s}^{t} - \frac{dt}{\tau_s}\,I_{s}^{t} + W_{\mathrm{in}}\,x^{t} + W_{\mathrm{rec}}\,s^{t} \label{eq:lif_syn} \\
  V_{m}^{t+1} &= V_{m}^{t} + \frac{dt}{\tau_m}\left(-V_{m}^{t} + I_{s}^{t} + I_{b}\right) \label{eq:lif_mem} \\
  s^{t} &= H\!\left(V_{m}^{t} - V_{\mathrm{th}}\right), \quad V_{m}^{t} \leftarrow V_{m}^{t} - V_{\mathrm{th}} \cdot s^{t} \label{eq:lif_spike}
\end{align}
where $\tau_s$ and $\tau_m$ are the synaptic and membrane time constants, $x^t$ is the input spike vector, $s^t$ is the binary output spike vector, $W_{\mathrm{in}}$ and $W_{\mathrm{rec}}$ are the input and recurrent weight matrices, $I_b$ is the bias current, $V_{\mathrm{th}}$ is the firing threshold, $H(\cdot)$ is the Heaviside step function, and $dt$ is the simulation timestep. The agent learns to maximize the expected return using the Q-learning update rule:
\begin{equation}
  Q(s_t, a_t) \leftarrow Q(s_t, a_t) + \eta \left[ r_{t+1} + \gamma \max_{a'} Q(s_{t+1}, a') - Q(s_t, a_t) \right]
\end{equation}
where $Q(s_t, a_t)$ is the action-value function for state $s_t$ and action $a_t$, $\eta$ is the learning rate, $r_{t+1}$ is the reward received at the next time step, and $\gamma \in [0, 1)$ is the discount factor.

The teacher DSQN observes the 80-channel spike matrix from the CBGT model (8 neuron populations $\times$ 10 neurons each), which is reduced to 16 channels by a fixed, parameter-free average-pooling stage (Section~\ref{sec:dimred}) before entering the trainable network; the teacher and all distilled students therefore operate on an identical 16-channel input. The trainable network consists of two hidden layers of 128 LIF neurons each, with membrane time constant $\tau_m = 100$~ms, synaptic time constant $\tau_s = 50$~ms, firing threshold $V_{\mathrm{th}} = 1.0$, and simulation time step $dt = 10$~ms. The output layer contains 9 LIF neurons corresponding to 9 discrete actions. Each observation spans 100 time steps (1~s of simulated neural activity), giving the recurrent dynamics sufficient temporal context for state estimation.

\subsection{Action Decoding via Partitioned Population Coding}
We decode actions via partitioned population coding to translate high-dimensional neural activity into low-dimensional clinical controls. The SNN output layer ($N_{out}=9$) is partitioned into three distinct sub-populations, each dedicated to a specific stimulation parameter: frequency ($f$), pulse width ($\delta$), and amplitude ($A$). We utilize a rate-based WTA readout where the network estimates Q-values by accumulating membrane potentials over the simulation window $T_{win}$. For each parameter $k \in \{f, \delta, A\}$, the discrete action $a_k \in \{-1, 0, +1\}$ (corresponding to decrease, maintain, or increase) is determined by the neuron with the maximum aggregate activity:
\begin{equation}
  a_k = \underset{j \in \{0,1,2\}}{\mathrm{argmax}} \left( \sum_{t=1}^{T_{win}} U_{k,j}[t] \right) - 1
\end{equation}
where $U_{k,j}[t]$ represents the membrane potential of the $j$-th neuron in the $k$-th sub-population at time $t$. This decoding scheme creates a reliable interface between the stochastic spiking domain and the deterministic control required by the DBS controller.

\subsection{Action Space and Clinical Constraints}
The policy outputs actions $a_t \in \mathbb{R}^3$ corresponding to relative adjustments in $f$, $A$, and $\delta$. To ensure the learned policy remains clinically viable and deployable on low-power hardware, we imposed strict saturation limits on these parameters (Table~\ref{tab:constraints}).

\textbf{Frequency ($f$):} The upper bound of 180~Hz was selected to align with the standard therapeutic window for high-frequency stimulation in Parkinson's disease, which typically ranges from 130--185~Hz~\cite{kraussTechnologyDeepBrain2021,brockerOptimizedTemporalPattern2017}. This limit also matches baselines established in RL environments and model-based control studies for stabilizing thalamic activation~\cite{gaoModelBasedDesignClosed2020, kuzminaNeurophysiologicallyRealisticEnvironment2025}.

\textbf{Pulse Width ($\delta$):} We constrained the pulse width to the range 60--400~$\mu$s. The lower bound ensures effective neural depolarization consistent with the chronaxie of myelinated axons, while the upper bound preserves battery life~\cite{kuncelSelectionStimulusParameters2004, sharmaStudyBatteryReplacement2023}.

\textbf{Amplitude ($A$):} The simulation implements current-controlled stimulation, which is consistent with the design of modern IPGs that regulate delivered charge directly~\cite{parastarfeizabadiAdvancesClosedloopDeep2017}. In this formulation, voltage is not an independently controllable parameter; it emerges implicitly from the product of the injected current and the tissue impedance at the electrode interface, and is therefore not constrained explicitly. We bounded the current amplitude to a maximum of 250~$\mu$A, which aligns with the operating envelope of next-generation ultra-low-power neuromorphic ICs designed for chronic implantation~\cite{parastarfeizabadiAdvancesClosedloopDeep2017} and serves as a safety limit against excessive charge density at the electrode-tissue interface.

We define an energy-aware penalty term $E$, based on the applied stimulation current, to quantify the energy consumption of the DBS policy. This term is calculated as the Root Mean Square (RMS) of the injected stimulation current waveform, $I_{dbs}$, over a given simulation window.
Formally, for a continuous stimulation waveform $I_{dbs}(t)$ over a time interval $T$, the energy proxy is defined as:
\begin{equation}
  E = \sqrt{ \frac{1}{T} \int_{0}^{T} \big( I_{dbs}(t) \big)^2 \,dt }
\end{equation}
In the context of our discrete-time biophysical simulation with a time step $\Delta t$ and a total of $N = \frac{T}{\Delta t}$ simulation steps, the RMS energy is computed empirically as:
\begin{equation}
  E = \sqrt{ \frac{1}{N} \sum_{i=1}^{N} \big( I_{dbs}[i] \big)^2 }
  \label{eq:rms_power}
\end{equation}
where $I_{dbs}[i]$ is the instantaneous stimulation current (in $\mu A / cm^2$) applied to the STN at the $i$-th time step. By integrating the squared amplitude over time, this formulation captures the interacting effects of all three tunable DBS parameters on cumulative power consumption.
\begin{table}[h]
  \centering
  \caption{Deep brain stimulation parameter constraints.}
  \label{tab:constraints}
  \small
  \begin{tabular*}{\textwidth}{@{\extracolsep{\fill}} l l l @{}}
    \toprule
    \textbf{Parameter} & \textbf{Range} & \textbf{Rationale} \\
    \midrule
    Frequency ($f$) & 0 -- 180 Hz & Upper bound of therapeutic window \\
    Pulse Width ($\delta$) & 60 -- 400 $\mu$s & Chronaxie limits and battery preservation \\
    Amplitude ($A$) & 0 -- 250 $\mu$A & Next-gen low-power integrated circuit limits \\
    \bottomrule
  \end{tabular*}
\end{table}

\subsection{Adaptive Reward Formulation}
The core strength of our approach lies in the co-optimization of therapeutic efficacy and stimulation energy. To achieve this, we formulate the DSQN reward function $r_t$ as a discontinuous function that prioritizes clinical symptom suppression above a defined $\beta$-power threshold ($\tau_\beta$), and energy conservation below it:
\begin{equation}
  r_t =
  \begin{cases}
    -\kappa \cdot (\beta_t - \tau_\beta) & \text{if } \beta_t > \tau_\beta \\
    \tau_{\mathrm{reward}} \cdot \left[ (1-\alpha) + \alpha \left( 1 - \min\left(\frac{E_t}{E_{\mathrm{max}}}, 1\right) \right) \right] & \text{if } \beta_t \leq \tau_\beta
  \end{cases}
\end{equation}
where $\beta_t$ is the instantaneous $\beta$-band power, $E_t$ is the instantaneous simulation RMS energy proxy (Eq.~\ref{eq:rms_power}), $E_{\mathrm{max}}$ is a normalization constant, $\tau_{\mathrm{reward}}$ is the base therapeutic reward, $\kappa$ is the suppression penalty coefficient, and $\alpha$ is a hyperparameter balancing the target outcome against the energy-savings gradient. The energy term enters the reward as the normalized ratio $E_t/E_{\mathrm{max}}$ rather than an absolute current. The policy therefore optimizes the fraction of charge it delivers relative to its own operating envelope, and the learned savings are a proportion of the baseline rather than a fixed quantity. $\kappa$ is matched to $\tau_{\mathrm{reward}}$ to establish proportionate incentives for symptom suppression and threshold violation; both are set large enough to keep temporal difference errors in the $L_1$ regime of the Huber loss throughout early training, stabilizing gradient updates before the policy discovers the therapeutic regime. This formulation guarantees the agent only receives positive reinforcement when the $\beta$-band pathological signature is sufficiently suppressed. We note that $\tau_\beta$ is an empirically derived fixed absolute threshold used to shape the training reward. The complete deep spiking Q-learning procedure is summarized in Algorithm~\ref{alg:energy_aware_rl}. Full implementation details, including the multi-head action structure, Q-value readout, and sparsity regularization, are provided in Algorithm~\ref{alg:ea_dsqn_detailed} (Appendix~\ref{app:eadsqn}).

\begin{algorithm}[t]
  \caption{Energy-Aware Deep Spiking Q-Learning (DSQN)}
  \label{alg:energy_aware_rl}
  \begin{algorithmic}[1]

    \Require Environment $\mathcal{E}$; policy SNN $\mathcal{S}_{\theta}$; target SNN $\mathcal{S}_{\theta'}$
    \Require Replay buffer $\mathcal{D}$; discount $\gamma$; soft-update coefficient $\tau_{\mathrm{tgt}}$; energy-balance weight $\alpha$

    \State Initialize $\theta$ randomly; $\theta' \gets \theta$

    \For{each episode}
    \State Reset $\mathcal{E}$ and membrane potentials; observe $\mathbf{s}_1$

    \For{$t = 1, \dots, T_{\max}$}

    \State Select $a_t$ via $\epsilon$-greedy policy over $\mathcal{S}_{\theta}(\mathbf{s}_t)$
    \State Execute $a_t$; observe $\mathbf{s}_{t+1}$, $\beta$-power $\beta_t$, energy $E_t$

    \If{$\beta_t > \tau_\beta$}
    \State $r_t \gets -\kappa \cdot (\beta_t - \tau_\beta)$ \Comment{Suppression penalty}
    \Else
    \State $r_t \gets \tau_{\mathrm{reward}} \cdot \bigl((1-\alpha) + \alpha\,(1 - E_t/E_{\max})\bigr)$ \Comment{Energy-savings reward}
    \EndIf

    \State Store $(\mathbf{s}_t, a_t, r_t, \mathbf{s}_{t+1})$ in $\mathcal{D}$

    \If{$|\mathcal{D}| \ge B$}
    \State Sample mini-batch from $\mathcal{D}$
    \State Compute TD targets: $y_i \gets r_i + \gamma \max_{a'} Q_{\theta'}(\mathbf{s}_{i+1}, a')$
    \State $\mathcal{L} \gets \mathrm{SmoothL1}\bigl(Q_\theta(\mathbf{s}_i, a_i),\; y_i\bigr) + \mathcal{L}_{\mathrm{sparsity}}$
    \State Update $\theta$ via AdamW on $\nabla_\theta \mathcal{L}$
    \State Soft-update: $\theta' \gets \tau_{\mathrm{tgt}}\,\theta + (1-\tau_{\mathrm{tgt}})\,\theta'$
    \EndIf

    \EndFor
    \EndFor

  \end{algorithmic}
\end{algorithm}

\subsection{Sparsity-Constrained Knowledge Distillation}
Knowledge Distillation (KD) is applied to transfer the policy from a dense teacher network to a streamlined student SNN, which reconciles the high-dimensional control policy with the strict resource constraints of the target hardware. We minimize the Kullback-Leibler (KL) divergence between the teacher's soft Q-values ($Q_T$) and the student's Q-values ($Q_S$) across all simulated states $s$ and actions $a$, scaled by a temperature parameter $T_{\mathrm{KD}}$:
\begin{equation}
  \mathcal{L}_{KD} = T_{\mathrm{KD}}^2 \sum_{a} \text{softmax}\left(\frac{Q_T(s,a)}{T_{\mathrm{KD}}}\right) \log \left( \frac{\text{softmax}(Q_T(s,a)/T_{\mathrm{KD}})}{\text{softmax}(Q_S(s,a)/T_{\mathrm{KD}})} \right)
\end{equation}
While KD effectively compresses the policy representation, it does not inherently minimize synaptic throughput. Since dynamic power consumption in neuromorphic architectures scales linearly with the rate of SynOps~\cite{barchiEnergyEfficientLowlatency2024}, we augment the optimization objective with an explicit sparsity penalty. This term regularizes the average firing rate of each hidden layer toward a target sparsity $\rho$ using a Bernoulli KL divergence:
\begin{equation}
  D_{KL}^{\text{sparse}}(\rho \| \hat{\rho}_l) = \rho \log \frac{\rho}{\hat{\rho}_l} + (1 - \rho) \log \frac{1 - \rho}{1 - \hat{\rho}_l}
\end{equation}
where $\hat{\rho}_l = \frac{1}{N_l T} \sum_{i,t} S_{l,i,t}$ is the average firing rate across all neurons $i$ and time steps $t$ in layer $l$, and $\rho \in (0,1)$ is the target sparsity (lower values encourage fewer spikes). This formulation encourages the actual firing rate distribution to match a sparse Bernoulli prior.
The total loss function combines the distillation loss and the weighted sparsity penalty across all hidden layers:
\begin{equation}
  \mathcal{L}_{total} = \mathcal{L}_{KD} + \lambda \sum_{l} D_{KL}^{\text{sparse}}(\rho \| \hat{\rho}_l)
  \label{eq:total_loss}
\end{equation}
where $\lambda$ is a hyperparameter governing the trade-off between task performance and metabolic cost. By varying $\lambda$ across training runs while holding $\rho$ constant for a target sparsity level, we directly control the network's operating point on the energy-efficacy Pareto front. Algorithm~\ref{alg:energy_aware_kd} (Appendix~\ref{app:kd}) summarizes the complete distillation procedure, including the sparsity warm-up schedule used to stabilize early training.

\subsection{Hardware Deployment and Benchmarking}
Our distilled control policy's compliance with the strict operating constraints of implantable bio-electronics was verified by deploying the SNN on the SynSense XyloAudio 3 neuromorphic processor. This validation serves to bridge the ``sim-to-real'' gap by confirming that the theoretical latency and throughput of the sparse policy align with physical hardware measurements~\cite{barchiEnergyEfficientLowlatency2024}. To contextualize the neuromorphic efficiency gains, we benchmarked the SNN against two independently trained baselines deployed on an NVIDIA Jetson Orin Nano (CUDA)~\cite{nvidiaJetsonNano2019}. The ANN baseline is a rate-coded feedforward network with a matched two-hidden-layer architecture (80 input channels, 128 hidden units, 9 output actions) that collapses the temporal dimension of the input via mean pooling before executing a single feedforward pass. The RNN baseline replaces the hidden layers with GRUs~\cite{choPropertiesNeuralMachine2014}, a recurrent architecture that maintains a learned hidden state across time steps, enabling it to process the full 100-step observation sequentially. Both baselines use the same DQN training objective and action space as the SNN, isolating the effect of the computing paradigm (event-driven spiking versus rate-coded feedforward versus recurrent gated processing) from architectural differences.

\subsubsection{Hardware-Compatible Dimensionality Reduction}
\label{sec:dimred}
Interfacing biological-scale simulations with ultra-low-power neuromorphic hardware runs into a mismatch in channel capacity. The biophysical simulation produces spiking outputs from $N=80$ independent neurons. However, the SynSense XyloAudio 3 supports a maximum of 16 input channels.
This mismatch was resolved by implementing a spatial average pooling layer, which reduces the state space dimensions from $C_{in}=80$ to $C_{out}=16$ using a non-overlapping strided pooling operation. For an input spike tensor $X \in \mathbb{R}^{B \times T \times C_{in}}$, the output $Y$ is calculated as:
\begin{equation}
  Y_{b,t,k} = \frac{1}{K} \sum_{i=1}^{K} X_{b,t, (k-1)K + i}
\end{equation}
where $b$ and $t$ index the batch and time dimensions respectively, $K=5$ is the kernel size and stride, and $k \in \{1, \dots, 16\}$ is the output channel index. This operation effectively aggregates the activity of clusters of 5 neighboring neurons into a single ``super-neuron'' channel. This specific downsampling strategy preserves the firing rate statistics of the nucleus sub-regions without introducing complex computational overhead, ensuring the signal chain remains strictly power-efficient and suitable for edge deployment. The teacher and every distilled student are trained and evaluated on the identical 16-channel representation.

\subsubsection{Architecture and Deployment Pipeline}
The XyloAudio 3 is a mixed-signal asynchronous processor designed for sub-milliwatt, always-on processing~\cite{bosSubmWNeuromorphicSNN2022}. Our implementation utilizes the device's digital SNN core to process direct spike injection. Because the core operates on an event-driven basis, the instantiated LIF neuron states are updated only upon the arrival of input spikes, minimizing dynamic power consumption during periods of sparse neural activity.

To reconcile high-level simulation with low-level hardware constraints, we utilized the Rockpool library~\cite{muir_dylan_2019_4639684}. The Xylo architecture relies on 8-bit integer precision for both synaptic weights and membrane potentials. Accordingly, we implemented a quantization-aware training pipeline using Rockpool's surrogate gradient layers, which account for truncation effects during training to prevent performance degradation upon deployment. Leveraging standardized intermediate representations~\cite{pedersenNeuromorphicIntermediateRepresentation2024}, the trained computation graph is mapped to the physical resources of the Xylo core, optimizing the allocation of neuron blocks and synaptic memory. Finally, the mapped network is converted into a binary configuration bitstream and uploaded via the \textit{Samna} runtime interface to the hardware, where the core clock frequency can be configured to optimally balance inference throughput and power consumption.

\subsubsection{Benchmarking Protocol}
To quantify the advantage of the neuromorphic approach, we executed an offline benchmarking protocol using identical synthetic spike train inputs of 100 time steps with 10~ms resolution across all models. For the Xylo SNN, raw spiking activity sequences from the biophysical simulation were aggregated via spatial pooling and transmitted to the hardware input buffer as Address Event Representation packets; the core was operated at 6.25~MHz and power was sampled at 20~Hz via the on-chip monitoring interface. Power on the Jetson baselines was recorded using \textit{tegrastats} at 100~ms intervals during inference. Energy per inference is the product of mean inference power and execution latency, consistent with recent benchmarking protocols for efficient temporal architectures on edge hardware~\cite{pierroAcceleratingLinearRecurrent2025}.

\subsubsection{Stimulation Power Accounting}
The power a stimulator delivers to tissue differs from the power it draws from the implant, and only the latter governs battery life (Table~\ref{tab:power_budget}). The standard TEED metric~\cite{kossCalculatingTotalElectrical2005} gives the tissue-delivered power $P_{\mathrm{tissue}} = (V^2 f\,\delta)/Z$, but this accounts only for the resistive dissipation at the electrode--tissue interface. Because the current that delivers this charge is sourced from a compliance-voltage rail held well above the tissue voltage drop, the driver dissipates the difference. Measured current-driver efficiencies for implantable stimulators fall around $\eta_{\mathrm{driver}} \approx 0.36$--$0.48$~\cite{palomeque-mangutFullyIntegratedPowerEfficient2022}. We therefore recover the device-drawn power as $P_{\mathrm{drawn}} = P_{\mathrm{tissue}}/\eta_{\mathrm{driver}}$, taking $\eta_{\mathrm{driver}} = 0.45$ from within this measured range. We account only for stimulation and inference, the two costs the controller architecture and energy-aware policy actually govern. A fixed device-overhead term such as telemetry, sensing, and clocking~\cite{saricaImplantablePulseGenerators2021} would add equally across all conditions and only dilute the relative percentage of power saved. Omitting it is therefore conservative and leaves relative comparisons unchanged. Furthermore, because the rodent-scale currents used in simulation ($\leq$250~$\mu$A) understate clinical charge delivery by roughly an order of magnitude, we anchor the absolute values to a clinical cDBS reference (130~Hz, 90~$\mu$s, 3.5~V, $Z = 1$~k$\Omega$). Each controller's measured charge reduction relative to cDBS (Fig.~\ref{fig:energy_comparison}) then scales this baseline while remaining independent of the specific device boundary.

\section{Results}\label{sec2}
\subsection{Therapeutic Efficacy in the Pathological State}
We trained an SNN controller, rewarded jointly for (1) suppressing pathological $\alpha$-$\beta$ oscillations and (2) minimizing neurostimulation energy, in closed loop with a dopamine-depleted biophysical model of the CBGT circuit. We separately deployed a compressed version of this policy on the SynSense XyloAudio~3 neuromorphic processor to measure its power and latency on hardware. The full pipeline is shown in Fig.~\ref{fig:system_overview}, with further detail in the Methods.

\begin{figure}[t!]
  \centering
  \includegraphics[width=\linewidth]{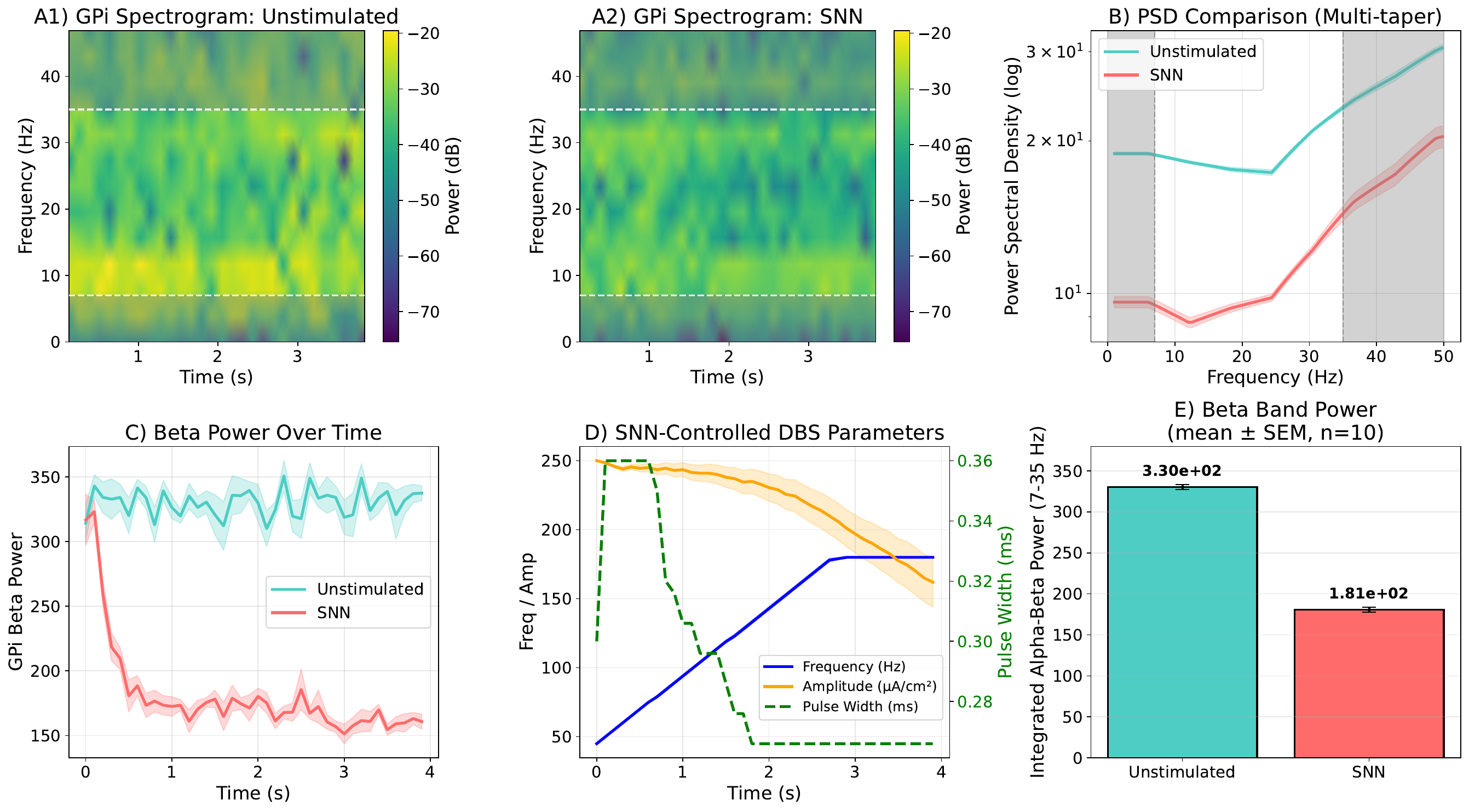}
  \caption{\textbf{Therapeutic efficacy of the neuromorphic controller.}
    \textbf{(A1--A2)}~Time-frequency spectrograms of GPi population spiking activity (short-time Fourier spectrogram, representative seed). The unstimulated state \textbf{(A1)} shows persistent high-power activation in the $\alpha$-$\beta$ band (7--35~Hz, white dashed lines); the SNN closed-loop state \textbf{(A2)} shows rapid suppression of this pathological synchronization.
    \textbf{(B)}~Multi-taper power spectral density (PSD) comparison between the unstimulated and SNN-controlled conditions. Shaded bands indicate the standard error of the mean (SEM) across $n=10$ seeds, and grey shading indicates frequencies outside the $\alpha$-$\beta$ band.
    \textbf{(C)}~Time course of $\alpha$-$\beta$ band power during the 4-second acute window; shaded bands are SEM ($n=10$).
    \textbf{(D)}~Adaptive stimulation parameters over the same window; shaded bands are SEM ($n=10$). Frequency (solid blue) ramps toward the 180~Hz ceiling; amplitude (solid orange) starts near 250~$\mu$A and gradually decreases over the stimulation window; pulse width (dashed green) exhibits greater variability across seeds.
  \textbf{(E)}~Mean integrated $\alpha$-$\beta$ band power for the PD baseline and SNN controller (mean $\pm$ SEM, $n=10$ seeds). Panels B and E use a multi-taper point-process estimator over per-100~ms windows, consistent with the online biomarker and the other experiments, while the A1--A2 spectrograms use a short-time Fourier estimate for time-frequency resolution. The rise in spectral power toward higher frequencies reflects the point-process firing-rate floor rather than a distinct oscillatory peak, so the therapeutic effect is a broadband reduction in $\alpha$-$\beta$ band power rather than suppression of an isolated beta peak.}
  \label{fig:efficacy}
\end{figure}

We first validated the controller's ability to mitigate symptoms during the active disease state. In the dopamine-depleted biophysical model, the GPi population exhibits exaggerated $\alpha$-$\beta$ oscillations, the hallmark electrophysiological signature of parkinsonian akinesia (Fig.~\ref{fig:efficacy}A).

Across 10 independent seeds, the SNN controller reduces $\alpha$-$\beta$ band integrated power by $45.2\%$ (95\% CI [42.6, 47.7]; $n = 10$; range 40--50\% across seeds; Fig.~\ref{fig:efficacy}E). The reduction is large and highly significant relative to the unstimulated PD baseline (two-sided paired $t(9) = 31.9$, $p = \num{1.4e-10}$; Cohen's $d_z = 10.1$; Wilcoxon signed-rank $p = 0.002$). The RL agent rapidly detects the pathological state from GPi feedback and initiates high-frequency stimulation to break the recurrent synchrony between the STN and globus pallidus externus (GPe), flattening the power spectral density (PSD) curve in the 7--35~Hz range (Fig.~\ref{fig:efficacy}B).

The spectrogram comparisons (Fig.~\ref{fig:efficacy}A2) show that the controller restores a desynchronized firing regime comparable to the healthy baseline state. As shown in Fig.~\ref{fig:efficacy}D, the SNN controller displays a clear ramp-up policy. Starting from an initial stimulation state (40~Hz frequency, 250~$\mu$A amplitude, and 0.3~ms pulse width), the SNN quickly ramps frequency toward the 180~Hz ceiling while amplitude gradually decreases from its initial maximum. Pulse width varies modestly across seeds, consistent with its narrow three-step action range rather than a distinct policy preference. In this fixed dopamine-depleted state, with no healthy intervals for down-regulation, maximal-drive stimulation is the reward-optimal strategy. To verify that this decisive modulation is driven by dynamic, closed-loop observation rather than a collapsed open-loop policy, we conducted a sensory ablation study (Appendix~\ref{app:ablation}). The results demonstrate that the network actively tracks underlying physiological dynamics, reverting to a quiescent, zero-stimulation state when deprived of input spikes.

\subsection{Stimulation Energy Efficiency}
\label{sec:stim_energy}

\begin{figure}[t!]
  \centering
  \includegraphics[width=0.9\linewidth]{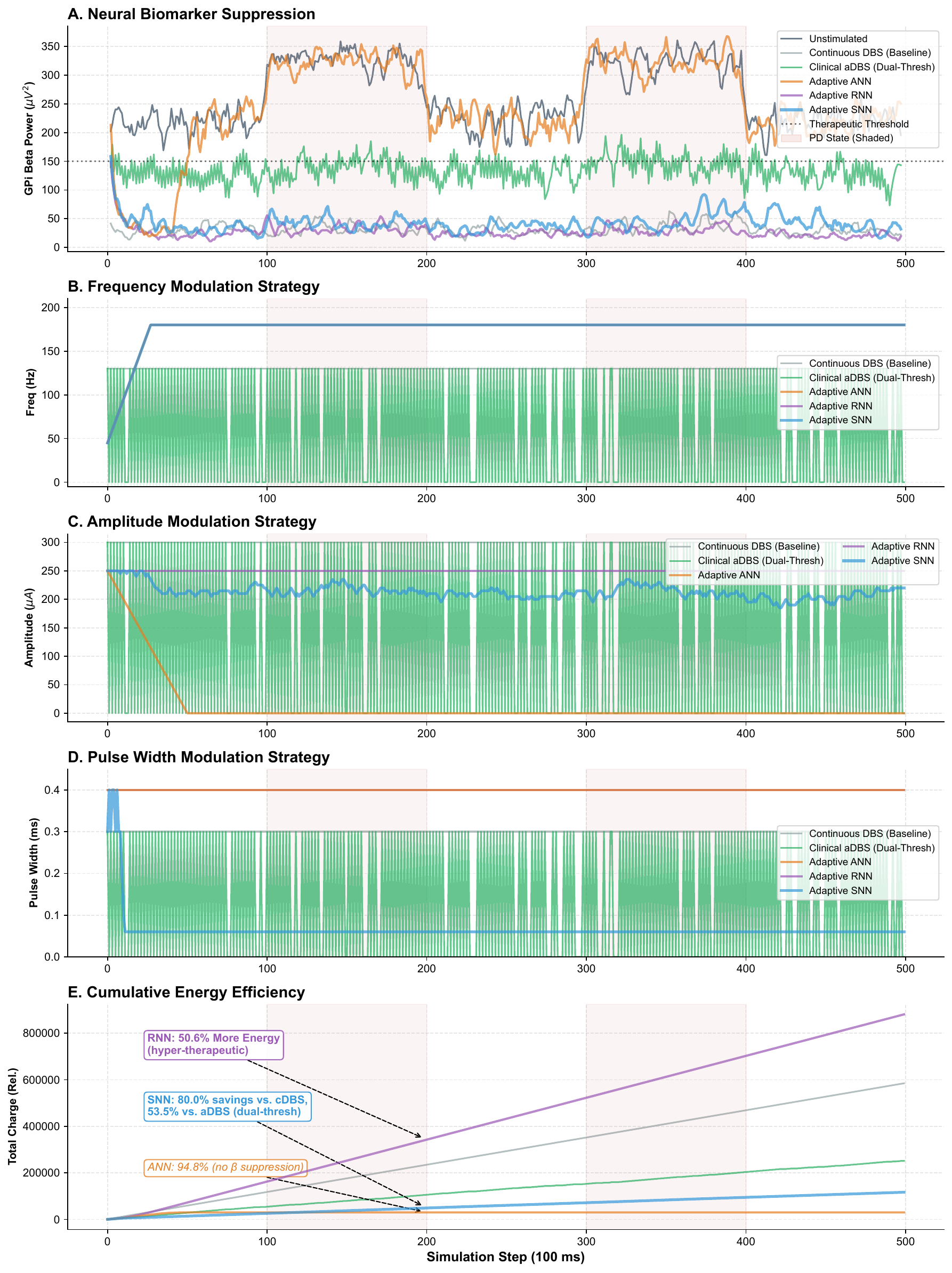}
  \caption{\textbf{Energy-aware stimulation comparison across alternating healthy and parkinsonian states.}
    \textbf{(A)}~GPi $\beta$-band power across six conditions: unstimulated (dark navy), continuous DBS (grey), clinical dual-threshold aDBS (green), ANN baseline (orange), RNN baseline (purple), and the adaptive SNN controller (blue). Light pink shading indicates parkinsonian (PD) blocks. The horizontal dotted line marks the therapeutic threshold ($\tau_\beta = 150~\mu V^2$).
    \textbf{(B--D)}~Stimulation parameter trajectories for frequency, amplitude, and pulse width.
  \textbf{(E)}~Cumulative charge delivery over the 50~s simulation. The ANN (94.8\% charge reduction, no $\beta$ suppression) and RNN (50.6\% more energy than cDBS, lack of adaptive down-regulation) each fail to discover an energy-aware policy, while the SNN learns an efficient modulation strategy.}
  \label{fig:energy_comparison}
\end{figure}

To quantify the stimulation energy savings of the energy-aware policy, we compared the SNN controller against conventional DBS strategies and two non-spiking baselines across alternating healthy and parkinsonian states (Fig.~\ref{fig:energy_comparison}). Six conditions were evaluated: no stimulation (pathological baseline), cDBS (130~Hz, 300~$\mu$A, 0.3~ms), a clinical dual-threshold aDBS heuristic that toggles stimulation based on $\beta$-power~\cite{littleAdaptiveDeepBrain2013, velisarDualThresholdNeural2019}, a feedforward ANN (Deep Q-Network, DQN) baseline, a recurrent RNN (Gated Recurrent Unit, GRU-DQN) baseline, and our adaptive SNN controller. The reward function, action space, and environment are held constant across all three learned controllers. The dual-threshold strategy represents the current clinical standard for adaptive DBS, in which stimulation is toggled on or off when a spectral biomarker crosses pre-set upper and lower bounds. This requires continuous online estimation of $\beta$-band power from the local field potential (LFP). The SNN controller, by contrast, receives only the raw multi-channel spike raster and therefore does not compute a spectral biomarker at any stage.

The SNN policy achieves an 80.0\% reduction in cumulative charge delivery relative to cDBS (Fig.~\ref{fig:energy_comparison}E), while maintaining $\beta$-power well below the therapeutic threshold during parkinsonian blocks. The SNN also outperforms the dual-threshold heuristic that represents the current clinical standard for adaptive DBS. In the cycling experiment, the dual-threshold aDBS heuristic cuts cumulative charge by only 57.0\% relative to cDBS, so the SNN delivers 53.5\% less charge than the heuristic. That larger saving costs nothing in efficacy: across the parkinsonian blocks of this experiment the SNN holds GPi $\beta$-power 85.9\% below the unstimulated baseline (mean \num{46.0}~$\mu V^2$), against 56.2\% for dual-threshold aDBS (mean \num{142.6}~$\mu V^2$, which by design tracks just under the $\tau_\beta = 150~\mu V^2$ threshold). Part of this margin is structural, since the heuristic toggles stimulation at fixed cDBS ON-parameters while the SNN can also modulate pulse width and amplitude, so a dual-threshold controller with an energy-optimized ON-state would narrow the gap. Even with that caveat, the SNN delivers both lower charge and deeper suppression than the heuristic. Rather than binary on/off switching, the SNN continuously modulates amplitude between $\sim$200--230~$\mu$A while maintaining frequency at 180~Hz and pulse width at 0.06~ms (Fig.~\ref{fig:energy_comparison}B--D). This narrow-pulse, sub-maximal amplitude strategy minimizes charge per pulse while preserving therapeutic neural recruitment (see Appendix~\ref{sec:sup_energy}). The cycling comparison is a single representative run of this alternating-state protocol, while the multi-seed statistical validation in this work covers the acute suppression result (Fig.~\ref{fig:efficacy}).

Neither non-spiking baseline produced a viable control policy. The ANN, operating on a mean-rate summary of the spiking input, converged to a degenerate strategy that minimizes amplitude to the floor (effectively not stimulating), achieving 94.8\% charge reduction but with no $\beta$ suppression (Fig.~\ref{fig:energy_comparison}A). The RNN, which processes the full 100-timestep observation through recurrent gates, failed to learn adaptive behavior, instead adopting a static, state-agnostic strategy that locks all three parameters at their maximum values (180~Hz, 0.4~ms, 250~$\mu$A) continuously regardless of physiological state. While this suppresses $\beta$-oscillations, it consumes 50.6\% \textit{more} energy than cDBS.



\subsection{Hardware-Efficient Compression via Sparsity Constraints}

The stimulation energy savings established above matter clinically only if the controller itself fits within an implant’s power budget. The teacher and every distilled student share the same input and topology, varying only in synaptic activity. The raw 80-channel CBGT output is spatially downsampled to 16 channels (applied identically in training and deployment, see Methods) and feeds an identical two-layer network. Trained with no constraint on firing, the teacher fires densely at 60.8~synaptic operations (SynOps; one weight accumulation per spike per outgoing synapse) per millisecond, above the power-efficient regime of the target hardware. Because dynamic power on neuromorphic cores scales with the rate of SynOps, the controller has to fire less without losing therapeutic efficacy. Finding that balance directly with RL is impractical. Each episode runs the biophysical simulation $\sim$50$\times$ slower than real time, which makes testing many sparsity settings prohibitively expensive. Distilling from the single fixed teacher avoids that cost, since regularizing each student toward a target firing rate lets us sweep dozens of sparsity configurations in hours rather than re-running RL from scratch for every one.

We integrated hardware constraints directly into the distillation objective by augmenting the loss function with a sparsity penalty term ($\lambda$, Eq.~\ref{eq:total_loss}). The student network learns to represent the therapeutic policy while exploiting activation sparsity, producing a ``hardware-efficient'' policy that minimizes the number of SynOps alongside its primary therapeutic objective.

\begin{figure}[t!]
  \centering
  \includegraphics[width=\linewidth]{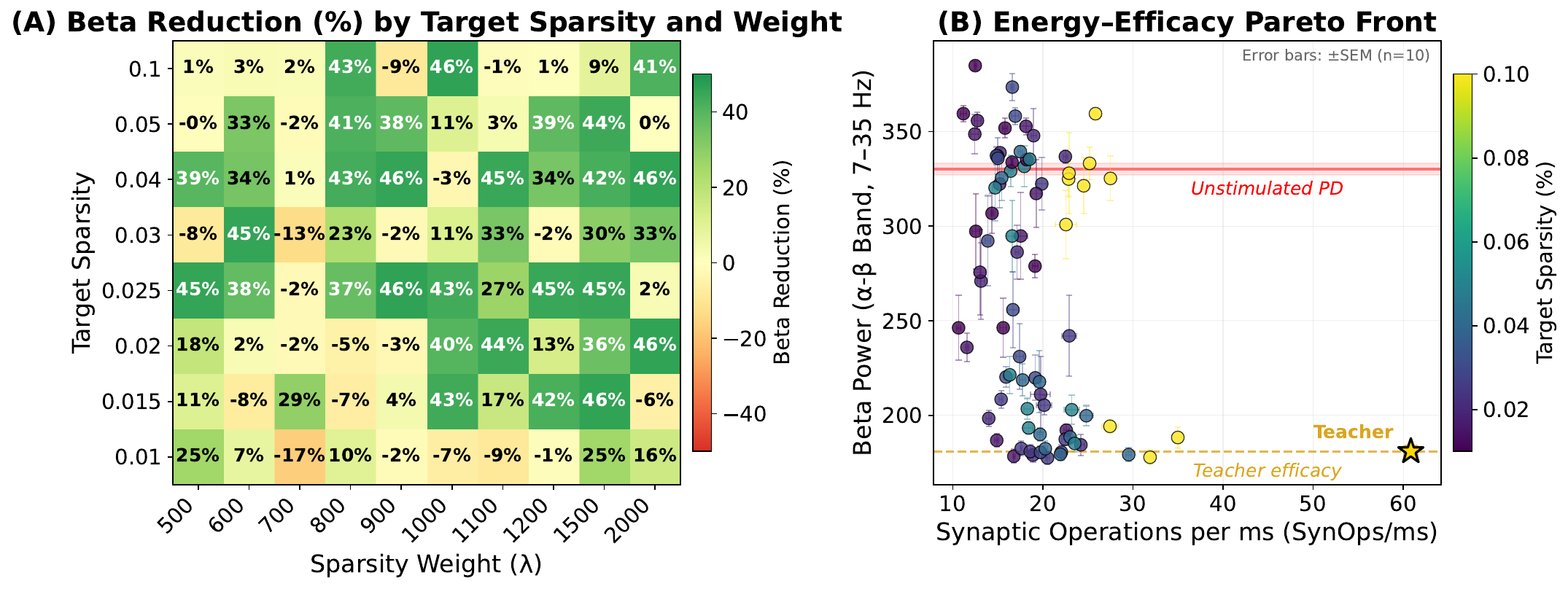}
  \caption{\textbf{Energy--Efficacy Trade-off in Sparsity-Constrained Distillation.}
  \textbf{(A)}~Heatmap of multi-taper $\alpha$-$\beta$ band power reduction (same metric as Fig.~\ref{fig:efficacy}; 10 seeds, 4-second window) across the $\rho \times \lambda$ hyperparameter grid (8~$\times$~10, 80 models). Increasing $\beta$ suppression transitions from red through yellow to green.
  \textbf{(B)}~Pareto front of SynOps/ms versus $\alpha$-$\beta$ band power (bottom-left is better). The red line marks the unstimulated PD baseline; models above it perform worse than no stimulation. The yellow dashed line marks teacher efficacy ($45.2\%$ at $60.8$~SynOps/ms). Of 80 students, 18 fall within 3 percentage points of the teacher at 1.7--4.1$\times$ fewer SynOps, while 21 fall above the PD baseline. The best Pareto-efficient student ($\rho=0.015$, $\lambda=1500$) reaches $45.9\%$ at $16.8$~SynOps/ms, $3.6\times$ fewer SynOps at equivalent efficacy.}
  \label{fig:pareto}
\end{figure}

We conducted a fine-grained hyperparameter sweep to systematically characterize the relationship between sparsity and therapeutic capacity. We trained 80 student policies across target sparsity levels ($\rho \in \{0.01, 0.015, 0.02, 0.025, 0.03, 0.04, 0.05, 0.10\}$) and penalty weights ($\lambda \in \{500, 600, 700, \ldots, 1500, 2000\}$) as depicted in Figure~\ref{fig:pareto}. The best Pareto-efficient student ($\rho=0.015$, $\lambda=1500$) matches teacher efficacy at $16.8$~SynOps/ms, $3.6\times$ fewer SynOps than the teacher (Fig.~\ref{fig:pareto}B). At the other end of the sweep, over-sparsified models provide no benefit over unstimulated PD. Higher penalty weight at moderate target sparsity consistently preserves efficacy, a pattern visible in the heatmap (Fig.~\ref{fig:pareto}A). Evaluated on the same multi-taper pipeline as the teacher (10 seeds, 4-second window), the best student suppresses $\alpha$-$\beta$ power by $45.9\%$ ($\pm 1.23\%$ SEM; $n=10$), statistically indistinguishable from the teacher's $45.2\%$ (paired $t(9) = -1.47$, $p = 0.176$; Wilcoxon $p = 0.232$). Compression therefore has minimal therapeutic impact, allowing the student to reach equivalent efficacy at $3.6\times$ fewer SynOps.

\subsection{Neuromorphic Hardware Efficiency}

Measuring the best student on the SynSense XyloAudio 3~\cite{bosSubmWNeuromorphicSNN2022} against equivalent baselines on the NVIDIA Jetson Orin Nano (CUDA) revealed a roughly 11,000$\times$ gap in power draw. This student achieves $45.9\%$ $\alpha$-$\beta$ suppression at $16.8$~SynOps/ms, matching teacher efficacy at $3.6\times$ lower synaptic throughput (see above).

\subsubsection{Power Consumption and Inference Latency}

\begin{figure}[htbp]
  \centering
  \includegraphics[width=\textwidth]{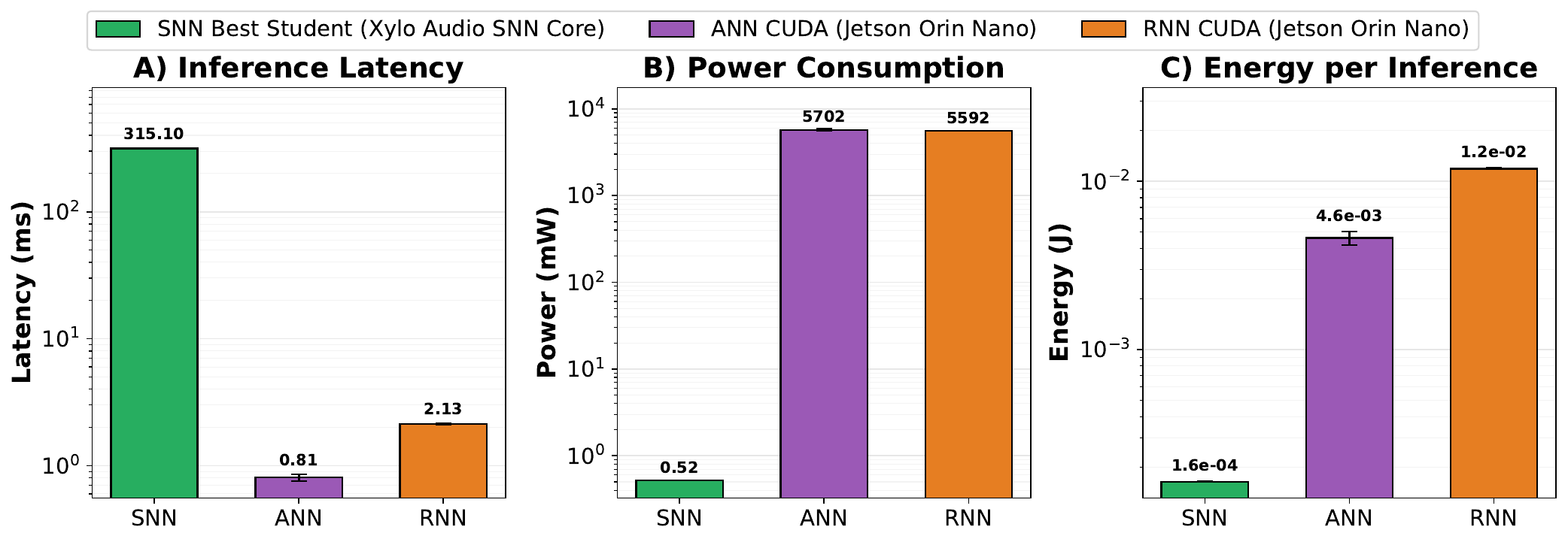}
  \caption{
    \textbf{Hardware benchmark comparison of neuromorphic (Xylo Audio 3) versus edge AI (Jetson Orin Nano CUDA) implementations.}
    \textbf{(A)} Inference latency.
    \textbf{(B)} Power consumption.
    \textbf{(C)} Energy per inference.
    The Xylo-deployed SNN consumes 28.1$\times$ less energy per inference than the ANN and 72.5$\times$ less than the RNN on the Jetson, despite being $\sim$390$\times$ slower, because of its $\sim$11,000$\times$ lower power draw.
  }
  \label{fig:hardware}
\end{figure}

Figure~\ref{fig:hardware} presents the measured power consumption, latency, and energy per inference across all configurations. The Xylo SNN drew 0.52~mW of mean system inference power (including IO and analog front-end), compared to the Jetson Orin Nano’s 5.7~W (ANN) and 5.6~W (RNN) on the VDD\_IN rail. To contextualize this discrepancy for clinical applications, powering the Jetson module with a hypothetical 5~Wh non-rechargeable DBS battery would deplete the implant's entire capacity in under 53~minutes solely from processing overhead. The 0.52~mW inference power of the Xylo processor, by contrast, permits over one year of continuous, always-on processing ($\sim$9,600~hours, derived as 5~Wh~/~0.52~mW). These measurements reflect different system boundaries (a standalone neuromorphic chip versus a complete edge-compute module). While both processors would ultimately require integration with an implant's data-acquisition harness, the comparison illustrates the baseline power required to execute neural networks under these two computing paradigms.

The Xylo-deployed SNN took a mean of 315.10~ms per inference, against 0.81~ms (ANN) and 2.13~ms (RNN) on the Jetson CUDA baselines, trading speed for its power efficiency. The RNN's longer latency relative to the ANN reflects the sequential nature of GRU recurrence, which cannot be fully parallelized on a GPU. The 315~ms inference latency is modest relative to the 1~s observation window from which each state is computed (100 steps $\times$ 10~ms/step), giving a total loop delay of $\sim$1.3~s per decision. This is consistent with the timescale over which $\beta$-band power envelopes vary in the parkinsonian state~\cite{littleAdaptiveDeepBrain2013} and with published adaptive DBS systems that update stimulation on second-to-sub-second timescales~\cite{velisarDualThresholdNeural2019}.

\subsubsection{Energy per Inference}

Energy per inference is the product of mean inference power and execution latency. Accumulated over the lifetime of a continuously running controller, it is what drains the battery. The Xylo student model consumed $1.64 \times 10^{-4}$~J per inference, compared to $4.61 \times 10^{-3}$~J for the ANN and $1.19 \times 10^{-2}$~J for the RNN on the Jetson CUDA, a 28.1$\times$ and 72.5$\times$ reduction respectively. The RNN's higher energy reflects both its longer sequential inference and the Jetson's sustained watts-level power draw. As with the power comparison, these ratios set the Xylo core against the Jetson’s full system.

Event-driven operation also lets the Xylo processor scale its power dynamically, dropping to a low-power idle state ($< 100$~$\mu$W) when no input spikes arrive, unlike the Jetson, which clocks continuously regardless of input activity. For the sparse firing patterns characteristic of biological signals, this idle-state efficiency would further widen the gap in real-world deployment.

\subsubsection{Implant Power Budget}

The neural network controller is only one component of the overall system power. The neurostimulator itself consumes a non-negligible portion of the power budget and must likewise be accounted for. To assess whether the stimulation savings from Section~\ref{sec:stim_energy} translate into meaningful battery life gains, we compared the stimulation power drawn from the device against the measured inference power of each platform (Table~\ref{tab:power_budget}). With a Jetson-class controller, inference power exceeds even clinical-scale stimulation by four orders of magnitude, rendering the system infeasible no matter how efficiently stimulation is managed, which is what makes sub-milliwatt neuromorphic inference a necessary condition for implantability. Only within that regime do the two costs become comparable, with stimulation accounting for roughly 38\% of the Xylo system's power budget at the clinical cDBS reference. The SNN policy's 85.6\% reduction in stimulation power then compounds with the low inference power to cut the stimulation share below 10\% and bring the total system draw to $\sim$0.57~mW, roughly a year of always-on operation on a 5~Wh cell. The RNN fails to achieve the reduction, drawing 28.2\% \textit{more} power than cDBS.

\begin{table}[htbp]
  \centering
  \caption{\textbf{Projected implant power budget.} Stimulation is reported as power drawn from the device, $P_{\mathrm{drawn}} = P_{\mathrm{tissue}}/\eta_{\mathrm{driver}}$, where the tissue-delivered power $P_{\mathrm{tissue}}$ comes from TEED~\cite{kossCalculatingTotalElectrical2005} at a clinical cDBS reference (130~Hz, 90~$\mu$s, 3.5~V, $Z = 1$~k$\Omega$, giving $\approx$143~mW$\cdot\Omega$ at the lower end of the reported clinical range of 145--220~mW$\cdot\Omega$~\cite{helmersComparisonBatteryLife2018}) and the current-driver efficiency $\eta_{\mathrm{driver}} \approx 0.45$ is taken from measured implantable stimulators~\cite{palomeque-mangutFullyIntegratedPowerEfficient2022}. Each controller's learned charge profile then scales this baseline. Inference power is measured on-chip (Xylo) and at the system rail (Jetson), and battery life is projected for a 5~Wh non-rechargeable IPG. The absolute stimulation values are a clinical-scale projection, whereas the relative reduction is measured directly and does not depend on the device boundary.}
  \small
  \begin{tabular*}{\textwidth}{@{\extracolsep{\fill}} l c c c @{}}
    \hline
    \textbf{Component} & \textbf{Xylo SNN} & \textbf{Jetson ANN} & \textbf{Jetson RNN} \\
    \hline
    Stim.\ power, cDBS ref.\ (drawn)     & 0.32~mW & 0.32~mW & 0.32~mW \\
    Stim.\ power, learned policy (drawn) & 0.05~mW & ${\approx}0$~mW\textsuperscript{$\dagger$} & 0.41~mW \\
    Inference (measured)                 & 0.52~mW & 5,703~mW & 5,592~mW \\
    \hline
    \textbf{Total (learned policy + inference)} & \textbf{0.57~mW} & \textbf{5,703~mW} & \textbf{5,592~mW} \\
    \textbf{Stim.\ share at cDBS baseline}      & \textbf{38\%} & \textbf{$<$0.01\%} & \textbf{$<$0.01\%} \\
    \textbf{Stim.\ TEED reduction versus cDBS}  & \textbf{85.6\%} & \textbf{---} & \textbf{$-$28.2\%} \\
    \textbf{Projected battery life (5~Wh)}      & \textbf{$\sim$1.0~yr} & \textbf{$<$1~hr} & \textbf{$<$1~hr} \\
    \hline
  \end{tabular*}

  \vspace{4pt}
  \begin{minipage}{\textwidth}
    \footnotesize
    \textsuperscript{$\dagger$}ANN converged to near-zero stimulation (94.8\% charge reduction versus cDBS; no therapeutic suppression).\\
    cDBS reference: 130~Hz, 90~$\mu$s, 3.5~V, $Z = 1$~k$\Omega$, $\eta_{\mathrm{driver}} \approx 0.45$ (clinical-scale projection).\\
    Simulation used 130~Hz, 300~$\mu$A, 0.3~ms; see Methods.
  \end{minipage}
  \label{tab:power_budget}
\end{table}

\section{Discussion and Conclusion}\label{sec3}

Hardware efficiency is a learnable property of the neural controller, one the training objective can optimize directly. The finding goes beyond power-efficient inference---including stimulation energy in the reward causes the policy to discover charge-efficient strategies that a standard objective would not. Once the controller falls within the implant's thermal budget, stimulation becomes a first-order remaining cost, and only a reward signal that penalizes it will drive the agent to reduce it.

\textbf{Co-optimizing controller and actuator energy.} Cutting only one of the two power costs does not deliver the savings, as the projected power budget makes clear (Table~\ref{tab:power_budget}). Sub-milliwatt inference is a prerequisite. Without it, the controller alone exceeds the implant's entire power budget, and any stimulation savings are irrelevant. But low-power inference alone is not sufficient, because the stimulator is still a meaningful part of the power budget. This joint optimization distinguishes the present framework from approaches that optimize only stimulation protocols~\cite{choClosedLoopDeepBrain2024} or only inference efficiency in isolation, because the therapeutic energy savings are only realizable on hardware that does not offset them. The 28.1$\times$ lower energy per inference than the ANN Jetson baseline (Fig.~\ref{fig:hardware}C) shows the advantage holds on a per-inference energy basis, though it inherits the system-boundary asymmetry from the power comparison (Fig.~\ref{fig:hardware}B). Additionally, the Jetson is a general-purpose edge module rather than the most power-frugal conventional option; a low-power microcontroller running the quantized ANN would draw far less than watts, bounding the $\sim$11,000$\times$ gap against general-purpose edge AI rather than a purpose-built low-power baseline. Nevertheless, the watts-level draw of full-system AI accelerators is infeasible for chronic implants, where battery replacement surgery is a primary constraint~\cite{sharmaStudyBatteryReplacement2023, parastarfeizabadiAdvancesClosedloopDeep2017}.

\textbf{Energy--efficacy trade-off and distillation failures.} The Pareto structure in Fig.~\ref{fig:pareto}B follows directly from the step-function reward. A student that cannot sustain suppression across seeds accumulates no therapeutic reinforcement. With only the sparsity penalty shaping training, it converges toward minimal firing rates and stimulation delivery, providing no improvement over the unstimulated state. Over-sparsification thus produces the same degenerate outcome as the ANN, not a graded reduction in efficacy. Moreover, the sparsity penalty enables the successful compressions by forcing the student to spend its limited representational budget on reproducing the teacher policy. This suffices at moderate regularization; pushing it further would cause the student to lose the capacity to maintain suppression across the seed-to-seed variability.

\textbf{Generalizability.} The energy--efficacy trade-off identified here is not specific to Parkinson's disease (Fig.~\ref{fig:pareto}). Responsive neurostimulation systems for epilepsy~\cite{nairNineyearProspectiveEfficacy2020} and closed-loop spinal cord stimulators for chronic pain~\cite{mekhailLongtermSafetyEfficacy2020} face the same operational constraints: neural biomarker inputs, actuator parameters with safety bounds, and chronic implant power envelopes in the milliwatt range. In each case, stimulation charge delivery becomes the dominant actuator cost once inference is efficient, making the energy-aware reward formulation transferable to these settings. The sparsity-constrained distillation and neuromorphic deployment steps are likewise architecture-agnostic, requiring only a task-appropriate simulation environment to adapt. As clinical neuromodulation moves from rigid heuristics toward continuous, state-dependent control, hardware-algorithm co-design of this kind will be necessary to prevent the controller's power consumption from undermining the therapeutic energy savings it produces.

\textbf{Spiking dynamics as inductive bias.} Holding the reward, action space, and environment fixed across all three models, the three-way ablation (ANN, RNN, SNN) isolates temporal processing and spiking dynamics contributions to policy quality. What changes is architecture and how each model processes its input: mean-pooled 80-channel vector for the ANN, full 80-channel sequence for the RNN, and spatially-pooled 16-channel sequence for the SNN. Given only these differences, the two baselines failed in opposite directions. With within-window spike timing averaged away, the ANN collapsed to the amplitude floor and stopped stimulating. The RNN went the other way, pinning every parameter at its safety ceiling and never down-regulating (Fig.~\ref{fig:energy_comparison}A). The SNN was the only architecture that found the energy-aware policy, modulating amplitude while holding frequency and pulse width at their boundary values.

Because the RNN retains both full temporal structure and richer channel information yet still fails to find an efficient policy, the ANN's failure cannot be attributed to loss of spike timing alone. Temporal processing is necessary but not sufficient. The SNN's leaky integrate-and-fire dynamics impose a structural constraint absent from continuous-valued GRU gates. Activations are sparse and binary, with membrane-level state decay. These dynamics limit the representational capacity of the network in a way that prevents convergence to the brute-force local optimum that the RNN finds. In our setup the spiking mechanism appears to act as an implicit architectural regularizer, paralleling the role of the explicit sparsity constraint during distillation. Both prevent convergence to degenerate policies, but through different mechanisms (neuron dynamics versus loss-function penalty). 

\textbf{Clinical deployment feasibility.} The current clinical state of the art, such as the Medtronic Percept\textsuperscript{TM}~PC~\cite{stanslaskiSensingDataMethodology2024, bronte-stewartLongTermPersonalizedAdaptive2025, velisarDualThresholdNeural2019}, relies on rigid dual-threshold heuristics that reduce high-dimensional neural dynamics into slow, single-variable spectral averages. They also require patient-specific threshold calibration. Processing directly on the raw spike raster, the SNN has no threshold to tune and can instead learn non-linear, state-dependent policies that co-modulate frequency, amplitude, and pulse width simultaneously, a combinatorial problem that rigid heuristics struggle to navigate dynamically~\cite{liuNeuralNetworkBasedClosedLoop2020}. The Xylo processor's asynchronous, event-driven architecture executes these policies at an IPG-compatible sub-milliwatt footprint. Operating within the 2$^\circ$C tissue heating safety limit~\cite{parastarfeizabadiAdvancesClosedloopDeep2017} and projecting to over a year of continuous operation on a standard IPG battery, this resolves the hardware constraint that previously precluded neural network controllers from chronic implants. The neuromorphic core is slower per inference than the GPU baselines, but for amplitude-envelope aDBS the costs that bind are power and energy, not speed. That latency is also a design choice rather than an intrinsic limit, since the Xylo core ran at 6.25~MHz and the $\sim$390$\times$ gap to the Jetson could be traded back against power by raising the clock frequency.

\textbf{Limitations.} While our neuromorphic framework achieves closed-loop therapeutic control within projected implant power budgets, the biophysical simulation environment introduces several practical constraints. First, because our results are grounded in a rodent biophysical model, they represent an approximation of the human CBGT circuit, meaning translation to clinical settings will ultimately require validation against patient-specific models and \textit{in vivo} experimentation. The quantity we expect to carry across this gap is the fractional saving rather than the rodent-scale absolute, because the energy-aware reward optimizes normalized charge and a policy that lowers delivered charge by 80\% saves that proportion regardless of the absolute current. At clinical amplitudes the same fraction corresponds to a larger absolute saving, though the exact value in humans will depend on patient-specific dynamics. Second, the controller currently optimizes a single biomarker ($\beta$-band power). Incorporating multi-biomarker approaches~\cite{gilronLongtermWirelessStreaming2021} would increase input dimensionality and require larger SNN architectures, necessitating a re-evaluation against the hardware constraints identified here. Finally, the biophysical simulation's approximately 50$\times$ slower-than-real-time speed was the primary training bottleneck, limiting episode throughput and constraining systematic exploration of network architectures, reward formulations, and hyperparameter configurations. Nonetheless, these results demonstrate the utility of energy-aware learning for implantable neuromodulation, with pathological $\beta$ suppression as the demonstrative use case that simultaneously improves both controller efficiency and neurostimulator energy management.


\begin{ack}
No specific external funding was received for this work, and the authors declare no competing
interests. All data are generated programmatically by the biophysical CBGT simulation model;
no external datasets were used. Code to reproduce all experiments is available at
\url{https://github.com/howyoubinh/CL-DBS-RL}.
\end{ack}

\bibliographystyle{unsrtnat}
\bibliography{references}

\clearpage
\setcounter{figure}{0}
\setcounter{table}{0}
\setcounter{equation}{0}
\setcounter{algorithm}{0}
\renewcommand{\thefigure}{S\arabic{figure}}
\renewcommand{\thetable}{S\arabic{table}}
\renewcommand{\theequation}{S\arabic{equation}}
\renewcommand{\thealgorithm}{S\arabic{algorithm}}
\renewcommand{\theHfigure}{app.\arabic{figure}}
\renewcommand{\theHtable}{app.\arabic{table}}
\renewcommand{\theHequation}{app.\arabic{equation}}
\renewcommand{\theHalgorithm}{app.\arabic{algorithm}}
\appendix

\begin{center}
  {\LARGE\bf Appendix}
\end{center}

\section{Detailed Neuronal Dynamics and Gating Kinetics}\label{app:dynamics}
This section provides the complete set of algebraic equations governing the voltage-dependent gating variables for the biophysical network model described in the Main Text Methods.

\subsection{Subthalamic Nucleus (STN) Neurons}
The STN neuron model includes current contributions from sodium ($I_{Na}$), potassium ($I_K$), A-type potassium ($I_A$), L-type calcium ($I_L$), T-type calcium ($I_T$), and calcium-dependent potassium ($I_{CaK}$) channels. The steady-state activation/inactivation functions ($x_\infty$) and time constants ($\tau_x$) for a gating variable $x$ are defined below. Voltage $V$ is in mV, and $[Ca]$ represents intracellular calcium concentration.

\begin{itemize}
  \item \textbf{Sodium Current ($I_{Na}$):} $m^3 h$ kinetics
    \begin{equation}
      m_\infty(V) = \frac{1}{1 + \exp\left(\frac{-(V+40)}{8}\right)}, \quad \tau_m(V) = 0.2 + \frac{3}{1 + \exp\left(\frac{-(V+53)}{-0.7}\right)}
    \end{equation}
    \begin{equation}
      h_\infty(V) = \frac{1}{1 + \exp\left(\frac{V+45.5}{6.4}\right)}, \quad \tau_h(V) = \frac{24.5}{\exp\left(\frac{-(V+50)}{-15}\right) + \exp\left(\frac{-(V+50)}{16}\right)}
    \end{equation}

  \item \textbf{Potassium Current ($I_K$):} $n^4$ kinetics
    \begin{equation}
      n_\infty(V) = \frac{1}{1 + \exp\left(\frac{-(V+41)}{14}\right)}, \quad \tau_n(V) = \frac{11}{\exp\left(\frac{-(V+40)}{-40}\right) + \exp\left(\frac{-(V+40)}{50}\right)}
    \end{equation}

  \item \textbf{A-type Potassium Current ($I_A$):} $a^2 b$ kinetics
    \begin{equation}
      a_\infty(V) = \frac{1}{1 + \exp\left(\frac{-(V+45)}{14.7}\right)}, \quad \tau_a(V) = 1 + \frac{1}{1 + \exp\left(\frac{-(V+40)}{-0.5}\right)}
    \end{equation}
    \begin{equation}
      b_\infty(V) = \frac{1}{1 + \exp\left(\frac{V+90}{7.5}\right)}, \quad \tau_b(V) = \frac{200}{\exp\left(\frac{-(V+60)}{-30}\right) + \exp\left(\frac{-(V+40)}{10}\right)}
    \end{equation}

  \item \textbf{L-type Calcium Current ($I_L$):} $c^2 d_1 d_2$ kinetics
    \begin{equation}
      c_\infty(V) = \frac{1}{1 + \exp\left(\frac{-(V+30.6)}{5}\right)}, \quad \tau_c(V) = 45 + \frac{10}{\exp\left(\frac{-(V+27)}{-20}\right) + \exp\left(\frac{-(V+50)}{15}\right)}
    \end{equation}
    \begin{equation}
      d_{1\infty}(V) = \frac{1}{1 + \exp\left(\frac{V+60}{7.5}\right)}, \quad \tau_{d1}(V) = 400 + \frac{500}{\exp\left(\frac{-(V+40)}{-15}\right) + \exp\left(\frac{-(V+20)}{20}\right)}
    \end{equation}
    \begin{equation}
      d_{2\infty}([Ca]) = \frac{1}{1 + \exp\left(\frac{[Ca]-0.1}{0.02}\right)}, \quad \tau_{d2}(V) = 130 \quad (\text{constant})
    \end{equation}

  \item \textbf{T-type Calcium Current ($I_T$):} $p^2 q$ kinetics
    \begin{equation}
      p_\infty(V) = \frac{1}{1 + \exp\left(\frac{-(V+56)}{6.7}\right)}, \quad \tau_p(V) = 5 + \frac{0.33}{\exp\left(\frac{-(V+27)}{-10}\right) + \exp\left(\frac{-(V+102)}{15}\right)}
    \end{equation}
    \begin{equation}
      q_\infty(V) = \frac{1}{1 + \exp\left(\frac{V+85}{5.8}\right)}, \quad \tau_q(V) = \frac{400}{\exp\left(\frac{-(V+50)}{-15}\right) + \exp\left(\frac{-(V+50)}{16}\right)}
    \end{equation}

  \item \textbf{Calcium-dependent Potassium Current ($I_{CaK}$):} $r^2$ kinetics
    \begin{equation}
      r_\infty([Ca]) = \frac{1}{1 + \exp\left(\frac{-([Ca]-0.17)}{0.08}\right)}, \quad \tau_r(V) = 2 \quad (\text{constant})
    \end{equation}

  \item \textbf{Intracellular Calcium Dynamics ($[Ca]$):}
    \begin{equation}
      \frac{d[Ca]}{dt} = -\alpha (I_L + I_T) - k_{Ca}[Ca]
    \end{equation}
    where $\alpha = 1/(ZF) = 5.18 \times 10^{-6}$ is a conversion factor describing the ion flux per unit current ($Z = 2$ is the calcium ion valence and $F = 96{,}485$~C/mol is Faraday's constant), and the calcium extrusion rate is $k_{Ca} = 2 \times 10^{-3} \text{ ms}^{-1}$.
\end{itemize}

\subsection{Other Neuronal Populations} The neuronal characteristics and gating variables for the Globus Pallidus (GPe/GPi), Thalamus (Th), and Striatal (D1/D2) neurons are grounded in the established Hodgkin-Huxley style models originally developed by Rubin and Terman~\cite{rubinTermanHighFrequency2004}. However, the specific equations, parameters, and Cortical Izhikevich neuron implementations used in this simulation are adopted from the comprehensive rat cortico-basal ganglia-thalamic (CBGT) network model described by Kumaravelu et al.~\cite{kumaraveluBiophysicalModelCortexbasal2016}.

\subsection{Synaptic Dynamics}
Inter-population synapses are modeled with conductance-based kinetics. The synaptic current for a given connection is:
\begin{equation}
  I_{syn}(t) = g_{max} S(t) (V(t) - E_{rev})
\end{equation}
where $g_{max}$ is the maximal synaptic conductance, $E_{rev}$ is the synaptic reversal potential, and the synaptic gating variable $S(t)$ follows second-order kinetics with distinct rise ($\tau_r$) and decay ($\tau_d$) time constants. This introduces realistic temporal latencies (e.g., N-methyl-D-aspartate (NMDA) $\tau_{decay} \approx 67$--$90$~ms) that the neuromorphic controller must navigate.

\subsection{Multi-Taper Point-Process Spectral Analysis}
For offline validation and spectral analysis, we computed power spectral densities (PSDs) directly from the discrete GPi population spike times using the multi-taper point-process method. This approach evaluates spectral variance by averaging independent estimates obtained from a set of $K$ orthogonal data tapers evaluated at exact spike times. Evaluating the spectrum at frequency $f$ for a set of $N_{sp}$ spike times $\{t_j\}$ is defined as:
\begin{equation}
  J_k(f) = \sum_{j=1}^{N_{sp}} h_k(t_j) e^{-i 2\pi f t_j} - \frac{N_{sp}}{N \Delta t} H_k(f)
\end{equation}
\begin{equation}
  S(f) = \frac{1}{K} \sum_{k=1}^K \left| J_k(f) \right|^2
\end{equation}
where $h_k(t_j)$ represents the $k$-th orthogonal Slepian taper (discrete prolate spheroidal sequence) interpolated to the spike time $t_j$, and the second term of $J_k(f)$ centers the point process by subtracting the mean rate component, with $H_k(f)$ defining the Fourier transform of the taper over the total duration $N \Delta t$. We selected a time-bandwidth product of $NW=3$, which mathematically allows for a maximum of $K = 2NW - 1 = 5$ strongly concentrated Slepian tapers. By averaging across these $K=5$ independent tapers, this approach ensures robust, low-variance estimation of oscillatory power independent of synaptic filtering assumptions or discrete time binning artifacts.

\section{Network Parameters}\label{app:netparams}
The intrinsic parameters for each neuron type used in the simulation are listed in Table \ref{tab:intrinsic_params}.

\begin{table}[h]
  \centering
  \caption{Maximal conductances ($mS/cm^2$) and reversal potentials ($mV$) for the biophysical model.}
  \label{tab:intrinsic_params}
  \begin{tabular}{lccccc}
    \toprule
    \textbf{Parameter} & \textbf{Thalamus} & \textbf{STN} & \textbf{GPe} & \textbf{GPi} & \textbf{Striatum (D1/D2)} \\
    \midrule
    Leak Conductance ($g_{Leak}$) & 0.05 & 0.35 & 0.1 & 0.1 & 0.1 \\
    Leak Reversal ($E_{Leak}$) & -70 & -60 & -65 & -65 & -67 \\
    Sodium ($g_{Na}$) & 3.0 & 49.0 & 120.0 & 120.0 & 100.0 \\
    Potassium ($g_K$) & 5.0 & 57.0 & 30.0 & 30.0 & 80.0 \\
    A-Type Potassium ($g_A$) & -- & 5.0 & -- & -- & -- \\
    T-type Calcium ($g_T$) & 5.0 & 5.0 & 0.5 & 0.5 & -- \\
    L-type Calcium ($g_L$) & -- & 15.0 & -- & -- & -- \\
    High-Thresh Ca ($g_{Ca}$) & -- & -- & 0.15 & 0.15 & -- \\
    Ca-dependent K ($g_{CaK}$) & -- & 1.0 & -- & -- & -- \\
    AHP Conductance ($g_{AHP}$) & -- & -- & 10.0 & 10.0 & -- \\
    \bottomrule
  \end{tabular}
  \vspace{0.2cm}\\
  \footnotesize{AHP: afterhyperpolarization conductance.}
\end{table}

\begin{table}[h]
  \centering
  \caption{Synaptic parameters used in the network simulation. $\tau_{rise}$ and $\tau_{decay}$ govern the synaptic gating variable $S(t)$. Delays ($t_d$) represent axonal propagation time.}
  \label{tab:synaptic_params}
  \begin{tabular}{llccc}
    \toprule
    \textbf{Source} $\to$ \textbf{Target} & \textbf{Receptor} & \textbf{Delay ($t_d$ ms)} & \textbf{$\tau_{rise}$ (ms)} & \textbf{$\tau_{decay}$ (ms)} \\
    \midrule
    Cortex $\to$ STN & AMPA & 5.9 & 0.5 & 2.49 \\
    Cortex $\to$ STN & NMDA & 5.9 & 2.0 & 90.0 \\
    Cortex $\to$ Striatum & AMPA & 5.1 & -- & 5.0* \\
    STN $\to$ GPe & AMPA & 2.0 & 0.4 & 2.5 \\
    STN $\to$ GPe & NMDA & 2.0 & 2.0 & 67.0 \\
    STN $\to$ GPi & AMPA & 1.5 & -- & 5.0* \\
    GPe $\to$ STN & GABA & 4.0 & 0.4 & 7.7 \\
    GPe $\to$ GPi & GABA & 3.0 & -- & 5.0* \\
    GPe $\to$ GPe & GABA & 1.0 & -- & 5.0* \\
    GPi $\to$ Thalamus & GABA & 5.0 & -- & 5.0* \\
    Striatum (D2) $\to$ GPe & GABA & 5.0 & -- & 5.0* \\
    Striatum (D1) $\to$ GPi & GABA & 4.0 & -- & 5.0* \\
    Thalamus $\to$ Cortex & AMPA & 5.0 & -- & 5.0* \\
    \bottomrule
  \end{tabular}
  \vspace{0.2cm}
  \\
  \footnotesize{*Modeled using single exponential decay ($\alpha$-function) where $\tau_{decay}$ dominates dynamics. AMPA: $\alpha$-amino-3-hydroxy-5-methyl-4-isoxazolepropionic acid; NMDA: N-methyl-D-aspartate; GABA: $\gamma$-aminobutyric acid.}
\end{table}

\section{EA-DSQN: Full Implementation Details}\label{app:eadsqn}

This section expands Algorithm~1 from the Main Text, providing the complete implementation-level procedure required for exact reproducibility. The key additions relative to the main-text version are: (i) the 3-head action decoding over independent output sub-populations; (ii) explicit Q-value readout via membrane-potential accumulation; (iii) per-head Bellman targets; (iv) the sparsity regularization loss applied to the two hidden LIF layers; and (v) gradient clipping via \texttt{clip\_grad\_value\_(100)} rather than norm clipping.

\begin{algorithm}[h]
  \caption{EA-DSQN --- Full Implementation}
  \label{alg:ea_dsqn_detailed}
  \begin{algorithmic}[1]

    \Require Environment $\mathcal{E}$; policy SNN $\mathcal{S}_{\theta}$; target SNN $\mathcal{S}_{\theta'}$
    \Require Replay buffer $\mathcal{D}$; batch size $B$; discount $\gamma$; soft-update coefficient $\tau_{\mathrm{tgt}}$
    \Require Exploration $\epsilon$; $\beta$-threshold $\tau_\beta$; energy-balance weight $\alpha$; suppression penalty $\kappa$
    \Require Sparsity weight $\lambda$; target sparsity $\rho$

    \State Initialize $\theta$ randomly; $\theta' \gets \theta$
    \State Initialize AdamW optimizer with learning rate $\eta$

    \For{episode $= 1, \dots, N$}
    \State Reset $\mathcal{E}$; obtain initial state $\mathbf{s}_1$
    \State Reset all LIF membrane potentials in $\mathcal{S}_{\theta}$ and $\mathcal{S}_{\theta'}$

    \For{$t = 1, \dots, T_{\max}$}

    \Statex \hspace{2em}\textit{// Action selection (3-head $\epsilon$-greedy)}
    \State $\mathbf{U}^{S},\, \mathbf{Z}^{S} \gets \mathcal{S}_{\theta}(\mathbf{s}_t)$ \Comment{Forward pass; $\mathbf{Z}^{S}$ are hidden spike trains}
    \State $\hat{\mathbf{u}} \gets \sum_{k} \mathbf{U}^{S}_k \in \mathbb{R}^{9}$ \Comment{Accumulate membrane potentials over time steps}
    \State With prob.\ $\epsilon$ sample random $a_t$; \textbf{else}
    \Statex \hspace{2em} $a_t^{(j)} \gets \arg\max\, \hat{\mathbf{u}}_{[3j:3j+3]},\ j \in \{0,1,2\}$ \Comment{$j{=}0,1,2$: $f$, $\delta$, $A$}

    \Statex \hspace{2em}\textit{// Environment step}
    \State Execute $a_t$ in $\mathcal{E}$; observe $\mathbf{s}_{t+1}$, $\beta$-power $\beta_t$, energy $E_t$

    \Statex \hspace{2em}\textit{// Reward (adaptive)}
    \If{$\beta_t > \tau_\beta$}
    \State $r_t \gets -\kappa \cdot (\beta_t - \tau_\beta)$ \Comment{Clinical suppression penalty}
    \Else
    \State $\bar{E} \gets \min(E_t / E_{\max},\; 1.0)$
    \State $r_t \gets \tau_{\mathrm{reward}} \cdot \bigl((1 - \alpha) + \alpha\,(1 - \bar{E})\bigr)$ \Comment{Energy-savings gradient}
    \EndIf

    \State Store $(\mathbf{s}_t,\, a_t,\, r_t,\, \mathbf{s}_{t+1})$ in $\mathcal{D}$;\; $\mathbf{s}_t \gets \mathbf{s}_{t+1}$;\; decay $\epsilon$

    \Statex \hspace{2em}\textit{// Optimisation}
    \If{$|\mathcal{D}| \ge B$}
    \State Sample $\{(\mathbf{s}_i, a_i, r_i, \mathbf{s}_{i+1})\}_{i=1}^B \sim \mathcal{D}$

    \State $\mathbf{U}^{S},\, \mathbf{Z}^{S} \gets \mathcal{S}_{\theta}(\mathbf{s}_i)$ \Comment{Policy forward pass}
    \State $\hat{\mathbf{u}}_i \gets \sum_k \mathbf{U}^{S}_k \in \mathbb{R}^{B \times 9}$
    \State $\mathbf{q}_i \gets \hat{\mathbf{u}}_i\bigl[\,[0,3,6] + a_i^{(j)}\,\bigr] \in \mathbb{R}^{B \times 3}$ \Comment{Gather per-head Q-values}

    \State $\mathbf{U}^{T},\, \mathbf{Z}^{T} \gets \mathcal{S}_{\theta'}(\mathbf{s}_{i+1})$ \Comment{Target forward pass}
    \State $\hat{\mathbf{u}}^{T}_i \gets \sum_k \mathbf{U}^{T}_k \in \mathbb{R}^{B \times 9}$
    \State $\mathbf{y}_i \gets r_i + \gamma\,\max_{a'}\,\hat{\mathbf{u}}^{T}_{i,[3j:3j+3]},\quad j\in\{0,1,2\}$ \Comment{Per-head Bellman target $\in \mathbb{R}^{B \times 3}$}

    \State $\mathcal{L}_{\mathrm{TD}} \gets \frac{1}{B}\sum_{i=1}^B \mathrm{SmoothL1}(\mathbf{q}_i,\, \mathbf{y}_i)$

    \Statex \hspace{4em}\textit{// Sparsity regularization on hidden LIF layers}
    \State $\bar{z}^{(l)} \gets \mathrm{mean}(\mathbf{Z}^{S,(l)}),\quad l \in \{1, 2\}$
    \State $\mathcal{L}_{\mathrm{sp}} \gets \lambda \sum_{l=1}^{2} \mathrm{KL}\!\left(\rho \;\|\; \bar{z}^{(l)}\right)$

    \State $\mathcal{L} \gets \mathcal{L}_{\mathrm{TD}} + \mathcal{L}_{\mathrm{sp}}$
    \State $\theta \gets \textsc{AdamW}\!\left(\theta,\;\nabla_\theta \mathcal{L},\;\texttt{clip\_grad\_value\_(100)}\right)$
    \State $\theta' \gets \tau_{\mathrm{tgt}}\,\theta + (1 - \tau_{\mathrm{tgt}})\,\theta'$ \Comment{Soft target update}
    \EndIf

    \EndFor
    \EndFor
  \end{algorithmic}
\end{algorithm}

\paragraph{Note on gradient clipping.} The implementation uses \texttt{clip\_grad\_value\_(100)}, which clips each individual gradient element to $[-100, 100]$, rather than the more common \texttt{clip\_grad\_norm\_}. This choice prevents outlier gradients from destabilising early training when the SNN membrane potentials are randomly initialized.

\begin{table}[h]
  \centering
  \caption{EA-DSQN hyperparameters used in all reported experiments.}
  \label{tab:ea_dsqn_hparams}
  \begin{tabular}{lll}
    \toprule
    \textbf{Hyperparameter} & \textbf{Symbol} & \textbf{Value} \\
    \midrule
    Discount factor          & $\gamma$               & 0.99 \\
    Soft update coefficient  & $\tau_{\mathrm{tgt}}$  & 0.005 \\
    Learning rate            & $\eta$                 & $1\times10^{-3}$ \\
    Replay buffer size       & $|\mathcal{D}|_{\max}$ & 100{,}000 \\
    Mini-batch size          & $B$                    & 128 \\
    $\epsilon$ initial       & $\epsilon_0$           & 0.9 \\
    $\epsilon$ final         & $\epsilon_{\min}$      & 0.05 \\
    $\epsilon$ decay steps   & ---                    & 2{,}000 \\
    Energy-balance weight    & $\alpha$               & 0.5 \\
    Base therapeutic reward  & $\tau_{\mathrm{reward}}$ & 3000 \\
    Suppression penalty      & $\kappa$                  & 30 \\
    $\beta$-power threshold  & $\tau_\beta$           & 150 $\mu$V$^2$ \\
    Max stimulation energy   & $E_{\max}$             & 250.0 $\mu$A \\
    Sparsity weight          & $\lambda$              & 0.0 (teacher/baseline); grid-searched in KD \\
    Target sparsity          & $\rho$                 & 0.0 (teacher/baseline); grid-searched in KD \\
    Gradient clip value      & ---                    & 100 \\
    SNN time steps per obs.  & $T_{\mathrm{win}}$     & 100 \\
    \bottomrule
  \end{tabular}
\end{table}

\clearpage

\section{Sparsity-Constrained Knowledge Distillation Algorithm}\label{app:kd}

\begin{algorithm}[h]
  \caption{Sparsity-Constrained Knowledge Distillation for Hardware-Constrained SNNs}
  \label{alg:energy_aware_kd}
  \begin{algorithmic}[1]

    \Require Trained teacher SNN $\mathcal{T}_{\theta_T}$; environment $\mathcal{E}$
    \Require Temperature $T_{\mathrm{KD}}$; target firing rate $\rho^*$; sparsity weight $\lambda$; learning rate $\eta$; epochs $N$

    \State Initialize student SNN $\mathcal{S}_{\theta_S}$ with random weights
    \State Initialize Adam optimizer with learning rate $\eta$

    \For{$n = 1, \dots, N$}

    \State Reset $\mathcal{E}$; obtain raw spike train $\mathbf{X} \in \{0,1\}^{T \times C}$
    \State $\mathbf{x} \gets \Call{AvgPool}{\mathbf{X}}$ \Comment{Downsample $C{=}80 \to 16$ channels}

    \Statex
    \State $\mathbf{U}^T \gets \mathcal{T}_{\theta_T}(\mathbf{x})$ \Comment{Teacher forward pass (frozen)}
    \State $\mathbf{q}^T \gets \sum_{t} \mathbf{U}^T_t$ \Comment{Temporal integration $\to$ Q-values}

    \Statex
    \State $\mathbf{U}^S,\; \{\mathbf{s}^{(l)}\}_{l=1}^{L_h} \gets \mathcal{S}_{\theta_S}(\mathbf{x})$ \Comment{Student forward pass}
    \State $\mathbf{q}^S \gets \sum_{t} \mathbf{U}^S_t$

    \Statex
    \State $\hat{\mathbf{p}}^T \gets \mathrm{softmax}\!\bigl(\mathbf{q}^T / T_{\mathrm{KD}}\bigr)$ \Comment{Soft teacher targets}
    \State $\hat{\mathbf{p}}^S_{\log} \gets \log\text{-}\mathrm{softmax}\!\bigl(\mathbf{q}^S / T_{\mathrm{KD}}\bigr)$ \Comment{Student log-probabilities}
    \State $\mathcal{L}_{\mathrm{KD}} \gets T_{\mathrm{KD}}^{2}\, D_{\mathrm{KL}}\!\bigl(\hat{\mathbf{p}}^S_{\log} \,\|\, \hat{\mathbf{p}}^T\bigr)$ \Comment{Behaviour-matching loss}

    \Statex
    \For{$l = 1, \dots, L_h$} \Comment{Sparsity penalty over hidden LIF layers}
    \State $\bar{\rho}_l \gets \mathrm{mean}(\mathbf{s}^{(l)})$ \Comment{Empirical mean firing rate}
    \EndFor
    \State $\mathcal{L}_{\mathrm{sparse}} \gets \lambda \sum_{l=1}^{L_h} D_{\mathrm{KL}}^{\,\mathrm{Bern}}\!\bigl(\rho^* \,\|\, \bar{\rho}_l\bigr)$ \Comment{Hardware-efficiency penalty}

    \Statex
    \State $w \gets \min\!\bigl(1,\; 2n / N\bigr)$ \Comment{Warm-up: linear ramp over first 50\%}
    \State $\mathcal{L} \gets \mathcal{L}_{\mathrm{KD}} + w \cdot \mathcal{L}_{\mathrm{sparse}}$ \Comment{Total loss}

    \Statex
    \State $\theta_S \gets \Call{Adam}{\theta_S,\; \nabla_{\theta_S}\mathcal{L},\; \mathrm{clip\_norm}{=}1.0}$ \Comment{Update student weights}

    \EndFor

    \State \Return $\theta_S$

  \end{algorithmic}
\end{algorithm}

\clearpage

\section{Energy-Aware Stimulation Experimental Protocol}
\label{sec:sup_energy}

To quantify the stimulation energy savings of the energy-aware control policy, we conducted a cycling experiment comparing six conditions across alternating healthy and parkinsonian states. The simulation cycled through five 100-step blocks (10~s each): Healthy $\rightarrow$ PD $\rightarrow$ Healthy $\rightarrow$ PD $\rightarrow$ Healthy, for a total of 500 steps (50~s). Six conditions were evaluated:
\begin{enumerate}
  \item \textbf{Unstimulated:} No stimulation applied (pathological baseline).
  \item \textbf{Continuous DBS (cDBS):} Fixed stimulation at 130~Hz, 300~$\mu$A, 0.3~ms.
  \item \textbf{Clinical adaptive DBS (aDBS; Dual-Threshold):} A dual-threshold heuristic modeled after the clinical protocol validated with the Medtronic Activa PC neurostimulator. The heuristic uses instantaneous GPi $\beta$-band power as its sole biomarker and applies a hysteresis band to prevent rapid stimulation cycling: stimulation activates at clinical parameters (130~Hz, 300~$\mu$A, 0.3~ms) when $\beta$-power exceeds an upper threshold ($\tau_{\mathrm{upper}} = 160~\mu V^2$) and deactivates when it falls below a lower threshold ($\tau_{\mathrm{lower}} = 140~\mu V^2$). Between the two bounds, the previous stimulation state is held.
  \item \textbf{ANN Baseline (DQN):} A feedforward Deep Q-Network processing the spike matrix as a mean-pooled rate vector.
  \item \textbf{RNN Baseline (GRU-DQN):} A recurrent GRU network processing the full 100-timestep observation through gated recurrent units.
  \item \textbf{Adaptive SNN:} The energy-aware SNN controller, which continuously co-modulates all three stimulation parameters.
\end{enumerate}


\clearpage

\section{State-Dependent Sensory Ablation}\label{app:ablation}

To rule out the hypothesis that the Energy-Aware SNN controller converged to an open-loop strategy, we performed a sensory ablation study during continuous simulation. This experiment isolates the network's dynamically computed state-dependent actions from its ingrained structural baseline reactions by entirely severing the sensory input pathway.

\begin{figure}[htpb]
  \centering
  \includegraphics[width=\textwidth]{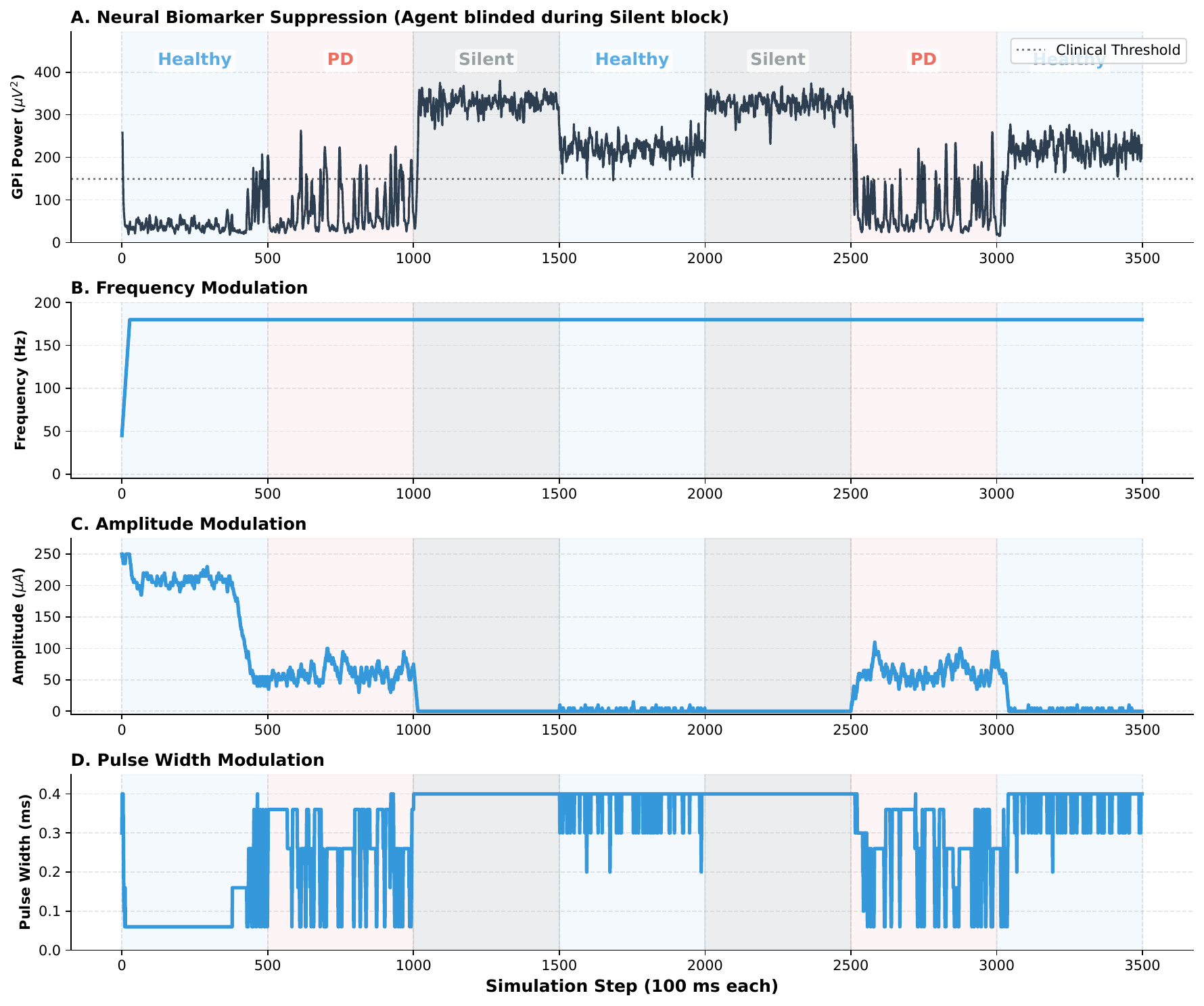}
  \caption{Sensory ablation study testing closed-loop behavior against a 3500-step sequence of Healthy, Parkinsonian, and Silent (sensory-deprived) states.}
  \label{fig:sup_real_to_silent}
\end{figure}

The SNN was evaluated across seven 500-step simulation blocks, subjecting the controller to a symmetrical sequence of environmental states and sensory deprivation (Figure~\ref{fig:sup_real_to_silent}):
\begin{enumerate}
  \item \textbf{Healthy (Block 1):} Natural physiological spiking from a healthy GPi.
  \item \textbf{Parkinsonian State (Block 2):} Closed-loop control against high $\beta$-band oscillations typical of Parkinson's disease.
  \item \textbf{Silent Brain (Block 3):} Sensory pathways are mathematically severed ($X=0$), simulating total sensor failure in an underlying pathological state.
  \item \textbf{Healthy Recovery (Block 4):} Sensory pathways are reconnected to healthy neural firing.
  \item \textbf{Silent Brain (Block 5):} Second sensory blackout period.
  \item \textbf{Parkinsonian Recovery (Block 6):} Immediate re-engagement against the pathological state.
  \item \textbf{Healthy (Block 7):} Final return to the healthy physiological baseline.
\end{enumerate}

During active closed-loop periods (Blocks 2 and 6), the SNN suppresses pathological GPi oscillations from a baseline of ${\sim}325~\mu V^2$ to $77.46~\mu V^2$ and $82.12~\mu V^2$, respectively, well below the clinical threshold of $150~\mu V^2$. These values reflect the dynamics of this specific alternating-state protocol and are not directly comparable to the acute efficacy measurements in the main text; the purpose here is to characterise how the SNN's learned dynamics are retained and reactivated across state transitions, not to replicate a specific therapeutic outcome.

When the simulation transitions to the Silent Brain blocks (Blocks 3 and 5), sensory input vectors drop to zero. The SNN's internal membrane potentials and synaptic currents decay naturally in the absence of input spikes. Rather than outputting chaotic or over-stimulating parameters, the SNN safely defaults to $0.00~\mu$A stimulation amplitude. Consequently, the untreated GPi network returns to its full pathological state (${\sim}320\text{--}325~\mu V^2$). This fail-safe behavior validates that the SNN acts as a pure reactive controller that behaves in a stable, predictable manner when blinded.

The controller exhibits high resilience and rapid re-engagement following sensory deprivation. At the transition from Block 5 to Block 6, immediately after spending 500 steps in a completely blind state with $0.00~\mu$A stimulation, the SNN is re-exposed to live pathological parkinsonian feedback. It instantly re-engages closed-loop control, rapidly ramping stimulation amplitude back to ${\sim}60~\mu$A and successfully restoring biomarker suppression. Finally, during the Healthy blocks directly following Silent blocks (Blocks 4 and 7), the mean local field potential (LFP) remains temporarily elevated (${\sim}210\text{--}220~\mu V^2$) compared to the initial Healthy baseline. This hysteresis occurs because the biophysical GPi population, left untreated in a highly pathological state during the silent blocks, establishes a persistent resonant oscillation that takes time to naturally decay after the environment switches back to a healthy state without active pathological inputs to drive the SNN.

\end{document}